\documentclass{article}
\usepackage[final]{colm2026_conference}

\usepackage{latexsym}

\usepackage[T1]{fontenc}

\usepackage[utf8]{inputenc}

\usepackage{microtype}
\usepackage{lineno}

\usepackage{inconsolata}

\usepackage{graphicx}

\usepackage[dvipsnames]{xcolor}
\definecolor{mycolor}{RGB}{38, 67, 166}
\usepackage{hyperref}
\hypersetup{colorlinks=true,citecolor=mycolor}

\usepackage{url}
\usepackage{booktabs}
\usepackage{amsfonts}
\usepackage{nicefrac}
\usepackage{colortbl}
\usepackage{multirow}

\usepackage{subcaption}
\usepackage{wrapfig}
\usepackage{adjustbox}
\usepackage{algorithm}
\usepackage{algorithmic}
\usepackage{amsmath}
\newcounter{observation}
\usepackage{placeins}

\usepackage{dblfloatfix}

\title{TRIM: Achieving Extreme Sparsity with Targeted Row-wise Iterative Metric-driven Pruning}

\author{
    Florentin Beck\thanks{Correspondence: \href{mailto:florentin.beck@uni-tuebingen.de}{florentin.beck@uni-tuebingen.de}}\\
    University of Tübingen\\
    School of Medicine
    \And
    William Rudman\\
    University of Texas at Austin\\
    Department of Linguistics
    \And
    Carsten Eickhoff\\
    University of Tübingen\\
    School of Medicine
}

\begin{document}
\raggedbottom
\ifcolmsubmission
\linenumbers
\fi
\maketitle
\lhead{\footnotesize Published at the COLM 2026 Workshop on Efficient Reasoning}
\begin{abstract}
Large language models (LLMs) present substantial computational and memory
challenges. Existing one-shot pruning methods often use a uniform sparsity
target for all rows within a layer, which can be too coarse at high sparsity.
This work introduces TRIM (\textbf{T}argeted \textbf{R}ow-wise \textbf{I}terative \textbf{M}etric-driven pruning), 
a forward-statistics-based wrapper that allocates different sparsity ratios to
individual output rows while keeping model weights frozen. TRIM separates the
choice of which weights to prune (scoring) from how many to prune in each row
(allocation), and composes with existing scorers and layer-wise allocations.
Across Qwen2.5, LLaMA-2, and OPT models, its largest gains occur at 70--80\%
sparsity. At 80\%, adding TRIM to OWL reduces WikiText-2 perplexity from 348.48
to 180.67 for Qwen2.5-14B and from 6461.43 to 324.14 for OPT-13B. 
We conclude that fine-grained, row-wise allocation is an effective mechanism for substantially improving one-shot pruning quality in the extreme-sparsity regime.
Code: \url{https://github.com/flobk/TRIM}.
\end{abstract}

\section{Introduction}
Large language models (LLMs) have gained prominence due to their generalized problem-solving abilities \citep{brown2020languagemodelsfewshotlearners, openai2024gpt4technicalreport, zhao2025surveylargelanguagemodels}.
This success is driven partly by model scaling \citep{kaplan2020scalinglawsneurallanguage, hoffmann2022trainingcomputeoptimallargelanguage}, which creates substantial deployment memory and compute demands \citep{zhou2024surveyefficientinferencelarge}.

Pruning reduces model size and computation by removing weights while seeking to preserve model quality \citep{Sekeletonization, OBD, han2015learningweightsconnectionsefficient}. 
Structured and semi-structured sparsity have historically been easier to accelerate, but their fixed patterns constrain which weights can be removed. 
Unstructured sparsity offers the finest-grained flexibility, allowing pruning methods to retain the weights most important to model quality. 
Recent GPU inference systems such as Flash-LLM and SpInfer demonstrate that specialized kernels can translate this flexibility into practical memory and speed gains \citep{xia2023flashllm,fan2025spinfer}. 
This makes preserving quality at high unstructured sparsity increasingly valuable, which is the regime targeted by TRIM.

Several pruning techniques have been developed to make LLMs more efficient. One major class induces sparsity during training or with fine-tuning, dynamically removing less important weights \citep{han2015learningweightsconnectionsefficient, evci2021rigginglotterymakingtickets, louizos2018learningsparseneuralnetworks}. 
While effective, these methods often require modifications to the training process and can be computationally expensive, particularly for models with billions of parameters, where maintaining optimizer states and gradients becomes a significant bottleneck \citep{molchanov2019importanceestimationneuralnetwork, liu2019rethinkingvaluenetworkpruning, hoefler2021sparsitydeeplearningpruning, gale2019statesparsitydeepneural}. 

As a result, one-shot pruning methods that reduce LLM parameters without retraining
to compute weight masks have gained prominence \citep{han2015learningweightsconnectionsefficient, sun2024simpleeffectivepruningapproach, zhang2024plugandplay, das2024sizegradientsshapepruning}.  
A challenge with current
approaches is that they often impose predefined structural constraints, such as uniform sparsity across all network layers or outputs. 
This imposed structure is arbitrary and can be detrimental to pruning performance, particularly as LLMs exhibit unique weight and activation characteristics, such as prominent outlier features and highly 
skewed activation distributions \citep{kovaleva-etal-2021-bert, dettmers2022llmint88bitmatrixmultiplication, rudman-etal-2023-outlier}. 

\begin{figure}[!t]
    \centering
    \includegraphics[
        width=0.97\linewidth,
        trim=4.25cm 9.8cm 3.99cm 10cm, 
        clip,
    ]{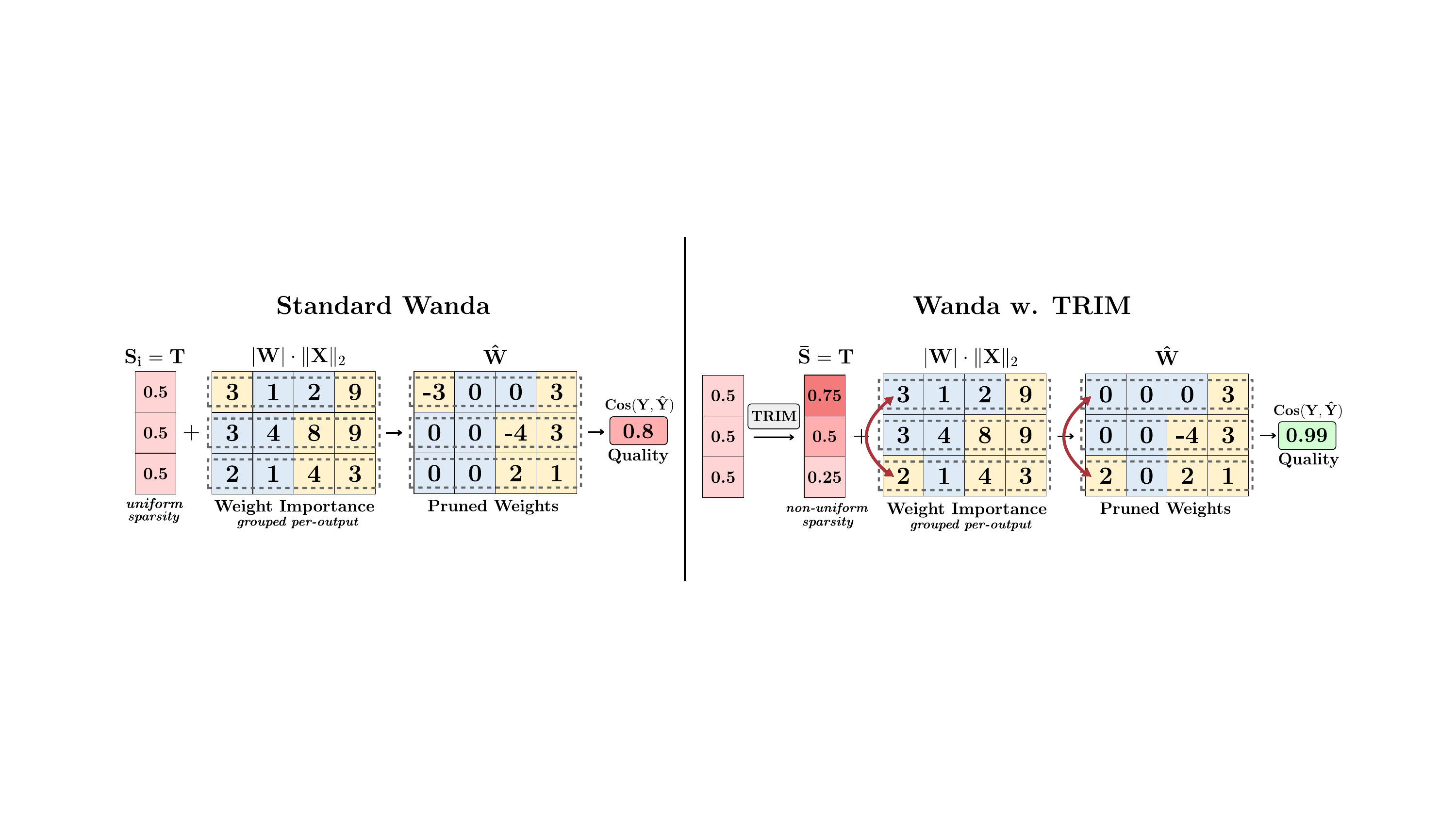}
    \caption{Illustrating non-uniform, dimension-wise sparsity. On the left is Wanda, which
    applies the layer sparsity ratio \(\mathbf{T}\) uniformly to all output dimension (rows) of the weight matrix \(\mathbf{W}\). TRIM iteratively defines sparsity ratios for individual dimensions in a non-uniform way. This targeted distribution of the available sparsity budget improves local (and global) pruning quality. The red arrows highlight weight positions whose pruning changes under TRIM's non-uniform row-wise allocation compared with uniform Wanda.}
    \label{fig:dimwise_sparsity_illustration}
\end{figure}
Layer-wise sparsity adaptation methods aim to mitigate this by assigning distinct sparsity ratios to each layer of the model, determined through algorithmic search \citep{li2024discovering, xu2025extremepruningllmsplugandplay}, 
heuristic-based approaches \citep{yin2024outlierweighedlayerwisesparsity, sun2025efficientshapleyvaluebasednonuniform}, or theoretically grounded strategies \citep{lu2024alphapruningusingheavytailedself}. While this offers more flexibility than uniform sparsity, these methods still apply a consistent sparsity target \textit{within} each layer and can struggle under extreme sparsity levels, such as removing more than 70\% of model parameters.

In this work, we propose TRIM: (\textbf{T}argeted \textbf{R}ow-wise \textbf{I}terative \textbf{M}etric-driven pruning). 
TRIM builds on the principles of one-shot pruning algorithms like Wanda \citep{sun2024simpleeffectivepruningapproach}, which compare and remove weights at each layer iteratively. Current methods typically apply a uniform sparsity constraint across all output dimensions \textit{(per-output)}, however, TRIM introduces a more granular approach. TRIM calculates unique sparsity ratios for \textit{individual output dimensions} (corresponding to a row in the layer's weight matrix).
This fine-grained allocation is guided by forward-pass quality statistics and an objective that often reduces variation in quality retention across outputs. We find the largest gains at extreme sparsity levels (70--80\% of weights removed).

Because TRIM adapts sparsity at the \textit{dimension-wise} level, it is inherently compatible with and can augment
existing \textit{layer-wise} allocation strategies. In this paper, we use TRIM to improve two layer-wise pruning methods: OWL and AlphaPruning. 

TRIM separates importance scoring, which orders weights for removal, from allocation, which sets each row's pruning budget. It therefore accepts different importance scorers and layer-wise allocations. The contributions of our work are as follows:

\begin{enumerate}
    \item We introduce TRIM, a lightweight and fast forward-only row-budget allocation mechanism that keeps weights frozen, is independent of the underlying importance scorer, and composes with layer-wise allocation.
    \item We evaluate TRIM across two layer allocations and four scorers, finding its largest improvements at 70--80\% sparsity; for example, OWL perplexity on OPT-13B at 80\% falls from \textit{6461} to \textit{324}.
    \item We conduct a thorough empirical analysis as to why TRIM leads to superior pruning results and show that individual output dimensions differ in their 1) sensitivity to pruning and 2) criticality for downstream performance.
\end{enumerate}

\section{Related Work}
A large body of work has explored pruning techniques that rely on gradient information or require retraining to recover performance. Gradient-based methods 
such as SNIP \citep{lee2019snipsingleshotnetworkpruning}, GraSP \citep{wang2020pickingwinningticketstraining}, and SynFlow \citep{tanaka2020pruningneuralnetworksdata} prune weights based on sensitivity estimates derived 
from gradients at initialization. Iterative methods like OBD \citep{OBD}, OBS \citep{OBS}, and movement pruning 
\citep{sanh2020movementpruningadaptivesparsity} leverage gradients or Hessians during or after training to inform pruning decisions, typically requiring retraining or fine-tuning to regain lost performance \citep{molchanov2019importanceestimationneuralnetwork, liu2019rethinkingvaluenetworkpruning, frankle2019lotterytickethypothesisfinding}.

While these methods achieve strong results, the computational burden of retraining models with billions of parameters has driven the development of efficient one-shot pruning approaches that eliminate the need for retraining. \cite{zhang2024dynamic, vanderouderaa2024llmsurgeon}

SparseGPT \citep{frantar2023sparsegptmassivelanguagemodels} applies one-shot pruning to LLMs and compensates for pruning errors by updating the remaining weights using approximate second-order information. Wanda \citep{sun2024simpleeffectivepruningapproach} offers a cheaper mask-only alternative without updating weights.
It addresses the limitations of magnitude pruning \citep{han2015learningweightsconnectionsefficient} (which fails due to LLMs having outlier activations) by incorporating input statistics of activations.
Wanda assigns a score to each weight based on both the absolute weight value and the L2 norm of a set of calibration inputs (\( |W| \cdot ||X||_2 \)).
Further, Wanda innovates the selection of the \textit{comparison group}, i.e., the subset of candidate weights from which the lowest-scoring weights are identified and removed to meet a sparsity target \(T\). Wanda adopts a \textit{per-output} (or row-wise) comparison, pruning the $T\%$ of lowest-scoring weights independently for each output dimension in the weight matrix.
This per-output strategy departs from prior approaches, like Magnitude pruning, which utilize the \textit{entire layer} as the comparison group.
However, the same per-output structure that contributes to Wanda's efficiency can also be a limitation, causing it to struggle at higher sparsity ratios compared to techniques with weight updates like SparseGPT \citep{frantar2023sparsegptmassivelanguagemodels}.

Methods such as Outlier Weighed Layerwise (OWL) sparsity extend Wanda by introducing \textit{layer-wise sparsity} ratios \citep{yin2024outlierweighedlayerwisesparsity}. OWL assigns unique sparsity
ratios to individual layers of a model while maintaining a global sparsity target. By assigning unique sparsity ratios to each layer, OWL avoids pruning layers that contain many 
\textit{outlier positions}. An outlier position exists when its Wanda score, $A_{ij}$, is greater than a predefined multiple, $M$, of the mean Wanda score of that layer, $\bar{A}$. Namely, if $A_{ij} > M \cdot \bar{A}$. AlphaPruning \citep{lu2024alphapruningusingheavytailedself} instead assigns layer-wise sparsity ratios based on the heavy-tailed shape of spectral densities in layer weight matrices.

AlphaPruning's allocation is motivated by heavy-tailed self-regularization theory \citep{martin2019rethinkinggeneralizationrequiresrevisiting, martin2019traditionalheavytailedselfregularization}, whereas OWL uses an outlier-based heuristic. Lua-LLM is the closest work at row-wise granularity: it learns layer- and intra-layer thresholds through gradient-based global optimization \citep{lu2025luallm}. TRIM instead wraps existing scorers and layer allocations, uses forward-pass calibration statistics, and keeps weights frozen. Broader end-to-end methods such as ELSA and LEAP optimize a global sparse model or per-weight masks \citep{lee2026elsa,mozaffari2026leap}; they use materially different, more expensive optimization regimes and are not matched substitutes in our experiments.

The evaluated Wanda, OWL, and AlphaPruning configurations assign a uniform sparsity ratio to every output row within a layer. This can be suboptimal at high sparsity because rows differ in pruning sensitivity. TRIM instead assigns row-specific ratios to \(W_{i,:}\), while retaining the adapted algorithm's original rule for ranking and selecting weights within each row.

\begin{algorithm*}[h]
\caption{Iterative Dimension-Wise Sparsity Adjustment}
\label{alg:dimwise_sparsity}
\begin{adjustbox}{max width=0.85\textwidth}
\begin{minipage}{1\textwidth}
\begin{algorithmic}[1]
    \STATE \textbf{Input:} Weight matrix \( W  \), Input activations \( X \), Target average sparsity \( T \), learning rate \( \alpha \)
    \STATE \textbf{Functions:} Prune(), Qmetric(), QmetricDimwise()
    \STATE \textbf{Output:} Selected dimension-wise sparsity vector \( S_{best} \).
    \STATE \textit{Compute unpruned output:} \( Y \leftarrow WX \)
    \STATE \textit{Start with uniform sparsity:} \( S_i \leftarrow T \) for \( i = 1, \dots, D \)
    \STATE \textit{Track best sparsity vector:} \( S_{best} \leftarrow S \)
    \STATE \textit{Track best quality score:} \( q_{best} \leftarrow -\infty \)
    \FOR{\( k = 0 \) to \( K-1 \)}
        \STATE \textit{Obtain pruned weights:} \( W_{pruned} \leftarrow\)  Prune(W, S)
        \STATE \textit{Compute pruned output:} \( \hat{Y} \leftarrow W_{pruned}X \)
        \STATE \textit{Calculate overall pruning quality for this iteration:} \( q_{k} \leftarrow \text{Qmetric}(Y, \hat{Y}) \)
        \IF{\( q_{k} > q_{best} \)} 
            \STATE \( q_{best} \leftarrow q_{k} \)
            \STATE \( S_{best} \leftarrow S \)
        \ENDIF
        \STATE \textit{Calculate quality per-output:} \( c_i \leftarrow \text{QmetricDimwise}(Y_{i,:}, \hat{Y}_{i,:}) \) for \( i = 1, \dots, D \)
        \STATE \textit{Normalize similarities to [0, 1]:} \( c'_i \leftarrow \frac{c_i - \min_{j} c_j}{\max_{j} c_j - \min_{j} c_j + \epsilon} \) for \( i = 1, \dots, D \)
        \STATE \textit{Apply learning rate:} \( \delta_i \leftarrow \alpha c'_i \) for \( i = 1, \dots, D \)
        \STATE \textit{Recenter mean:} \( S_i \leftarrow \delta_i - \frac{1}{D}\sum_{j=1}^{D} \delta_j + T \) for \( i = 1, \dots, D \)
    \ENDFOR
    \STATE \textit{Return selected sparsity allocation:} \textbf{Return} \( S_{best} \)
\end{algorithmic}
\end{minipage}
\end{adjustbox}
\end{algorithm*}

\section{TRIM}\label{sec:TRIM}

\begin{table}[t]
    \caption{Perplexity results on WikiText-2 for various models pruned at 70\% and 80\% sparsity. 
    The \(\Delta\) column shows the average perplexity reduction across the tested models achieved by adding TRIM. Lower is better.}
    \label{tab:perplexity_results}
    \centering
    \begin{adjustbox}{width=0.98\textwidth, center}
    \begin{tabular}{clcccccccccc}
        \toprule
        \multirow{2.5}{*}{ } & \multirow{2.5}{*}{Method} & \multicolumn{2}{c}{OPT} & \multicolumn{2}{c}{LLaMA-2}  & \multicolumn{5}{c}{Qwen2.5} \\ 
        \cmidrule(lr){3-4}
        \cmidrule(lr){5-6}
        \cmidrule(lr){7-11}
        & & 6.7B & 13B & 7B & 13B & 3B & 7B & 14B & 32B & 72B & \(\Delta\) \\
        \midrule
        0\% & \multicolumn{1}{c}{-} & 10.86 & 10.13 & 5.47 & 4.88 & 8.03 & 6.85 & 5.29 & 5.018 & 3.851 & - \\        
        \midrule
        \multirow{4}{*}{70\%}
        & Alpha    & 63.53 & 820.42 & 31.32 & 15.19 & 164.39 & 68.93 & 61.59 & 15.70 & 11.66 & 
        \multirow{2}{*}{\textit{-25.3}\%}\\
        & \cellcolor{gray!9}+TRIM &  \cellcolor{gray!9}\textbf{30.80} & \cellcolor{gray!9}\textbf{35.16} &  \cellcolor{gray!9}\textbf{29.00} & \cellcolor{gray!9}\textbf{14.70} & \cellcolor{gray!9}\textbf{128.44} & \cellcolor{gray!9}\textbf{62.22} & \cellcolor{gray!9}\textbf{46.43} & \cellcolor{gray!9}\textbf{13.94} & \cellcolor{gray!9}\textbf{11.41} & \\
        \cmidrule(l){2-12}
        & OWL       & 40.30& 28.92 & 21.20  & 16.02 & 83.98 & 44.60 & 36.46 & 13.06 & 11.25 & 
        \multirow{2}{*}{\textit{-4.6}\%} \\ 
        & \cellcolor{gray!9}+TRIM   &  \cellcolor{gray!9}\textbf{38.21} & \cellcolor{gray!9}\textbf{28.21} &   \cellcolor{gray!9}\textbf{20.10} & \cellcolor{gray!9}\textbf{14.39} & \cellcolor{gray!9}\textbf{83.39} & \cellcolor{gray!9}\textbf{43.97} & \cellcolor{gray!9}\textbf{33.21} & \cellcolor{gray!9}\textbf{12.28} & \cellcolor{gray!9}\textbf{11.06} \\
        \midrule
        \multirow{4}{*}{80\%}
        & Alpha     & 5357.60 & 4726.09 & 1558.41 & 143.19 & 9872.21 & \textbf{728.82} & 1040.91 & 469.52 & 70.26 &
        \multirow{2}{*}{\textit{-41.6}\%} \\
        & \cellcolor{gray!9}+TRIM & \cellcolor{gray!9}\textbf{656.26} & \cellcolor{gray!9}\textbf{557.30} &  \cellcolor{gray!9}\textbf{1527.01} & \cellcolor{gray!9}\textbf{108.70} & \cellcolor{gray!9}\textbf{1685.51} & \cellcolor{gray!9}1037.11 & \cellcolor{gray!9}\textbf{435.84} & \cellcolor{gray!9}\textbf{133.94} & \cellcolor{gray!9}\textbf{68.64}& \\
        \cmidrule(l){2-12}
        & OWL       & 8509.51 & 6461.43 &  \textbf{406.20} & 225.04 & 768.69 & 245.32 & 348.48 & 73.44 & 61.59 & 
        \multirow{2}{*}{\textit{-34.9}\%} \\
        & \cellcolor{gray!9}+TRIM & \cellcolor{gray!9}\textbf{492.47}  & \cellcolor{gray!9}\textbf{324.14} & \cellcolor{gray!9}{437.88} & \cellcolor{gray!9}\textbf{154.83} & \cellcolor{gray!9}\textbf{504.89} & \cellcolor{gray!9}\textbf{233.80} & \cellcolor{gray!9}\textbf{180.67} & \cellcolor{gray!9}\textbf{66.95} & \cellcolor{gray!9}\textbf{58.15} & \\ 
        \bottomrule
    \end{tabular}
    \end{adjustbox}
\end{table}

Figure \ref{fig:dimwise_sparsity_illustration} illustrates the core concept of TRIM. Let \( W \in \mathbb{R}^{D \times N} \) be the weight matrix of a layer, where \( D \) is the number of output dimensions (rows) and \( N \) is the number of input dimensions (columns). Each row \( W_{i,:} \) corresponds to the weights associated with the \(i\)-th output dimension.
We define a \textbf{dimension-wise sparsity vector} \( S = [S_1, S_2, \dots, S_D] \), where each element \( S_i \in [0, 1] \) specifies the target sparsity ratio for the \(i\)-th output dimension \(W_{i,:}\). This means that for each row \(W_{i,:}\), \(S_i \cdot N\) of its weights will be pruned. While the individual sparsity ratios \(S_i\) can vary, they must collectively satisfy the target sparsity \(T\) for the layer. This is enforced by ensuring their average equals \(T\):
\begin{equation} \label{eq:avg_sparsity_constraint}
\frac{1}{D} \sum_{i=1}^{D} S_i = T
\end{equation}
Setting \(S_i = T\) for all \(i \in \{1, \dots, D\}\) reverts to the standard pruning approach where sparsity is uniform across output dimensions.

The core challenge is to determine an effective allocation strategy for the dimension-wise sparsity vector, \(S\), that optimizes performance while adhering to Equation \ref{eq:avg_sparsity_constraint}. Below, we detail our method for finding \(S\), given a layer's weight matrix, \( W \), and \(L\) sample input activations \( X \in \mathbb{R}^{N \times L} \).

\paragraph{Calculating Dimension-Wise Sparsity Vectors.}
To find a suitable dimension-wise sparsity vector \( S \), we employ an iterative adjustment process detailed in Algorithm \ref{alg:dimwise_sparsity}. The process begins by computing the unpruned output \( Y \leftarrow WX \) (Line 4), with \( Y \in \mathbb{R}^{D \times L} \) and initializing \( S \) uniformly, so that \( S_i = T\) for all output dimensions \( i = 1, \dots, D \) (Line 5). This initial \( S \) also serves as the first candidate for the best sparsity vector, \( S_{best} \). The algorithm then iterates \( K \) times (Lines 8-20). In each iteration \( k \):
\begin{enumerate}
    \item The weight matrix \( W \) is pruned using the current sparsity vector to obtain \( W_{pruned}\) (Line 9). This means for each row \( W_{i,:} \), \( S_i \cdot N \) weights are removed based on the chosen pruning criterion (e.g., lowest Wanda scores).
    \item The pruned output \( \hat{Y} \leftarrow W_{pruned}X \) is computed (Line 10).
    \item The overall quality \( q_k \leftarrow \text{Qmetric}(Y, \hat{Y}) \) of the iteration is calculated (Line 11). If \( q_k \) is better than the best quality found so far (\( q_{best} \)), then \( q_{best} \) is updated, and the current \( S \) is stored as \( S_{best} \) (Lines 12-14). 
    \item The quality \( c_i \leftarrow \text{QmetricDimwise}(Y_{i,:}, \hat{Y}_{i,:}) \) for \textit{each individual output dimension} is calculated (Line 15). 
    \item The sparsity targets \( S_i \) are adjusted based on these per-dimension quality scores. 
    First, the scores \( c_i \) are normalized to a [0, 1] range (Line 16). An update term \( \delta_i \leftarrow \alpha c'_i \) is calculated using a learning rate \( \alpha \) (Line 17). This mechanism aims to increase the sparsity target \( S_i \) for dimensions exhibiting higher quality (less degradation, higher \( c'_i \)) and decrease it for dimensions with lower quality, thereby pruning less sensitive dimensions more aggressively.
    \item Finally, the adjusted sparsity vector is re-centered to ensure its average value remains equal to the overall target layer sparsity \( T \) (Line 18) in order to satisfy Equation \ref{eq:avg_sparsity_constraint}.
\end{enumerate}
 After \( K \) iterations, the algorithm returns \( S_{best} \) (Line 21), which represents the dimension-wise sparsity allocation that yields the highest overall quality \( q_{best} \) during the 
 iterative search. \( S_{best} \) is then used to perform the final pruning of the layer. In Appendix \ref{sec:sensitivity_k}, we explore the impact of $K$ on the pruning quality of TRIM. 

To determine the sparsity distribution, we evaluate positive learning rates \( \alpha \), beginning with small values, applying Algorithm~\ref{alg:dimwise_sparsity}, and recording \( q_{best} \). We increase \( \alpha \) and repeat the process, stopping when a larger value no longer improves \( q_{best} \). 
However, if a positive $\alpha$ does not yield a quality $q_{best}$ superior to the uniform sparsity baseline, we use
a negative learning rate. This allows for better generalization across different model activation spaces. In Section \ref{sec:Analysis},
we briefly explore when a negative learning rate occurs in models.

 We use cosine similarity as our quality metric and refer the reader to Appendix \ref{quality_metric} 
 for the evaluation of other metrics. Additionally, Appendix~\ref{runtime_overhead} presents our analysis of the computational overhead of this iterative procedure. 
 Our findings show that calculating the sparsity vectors for all layers of the Qwen2.5-14B model adds only a small delay, taking between 27.76 and 59.88 seconds based on the number of calibration samples used.

\begin{figure}[H]
    \centering
    \includegraphics[width=0.60\linewidth]{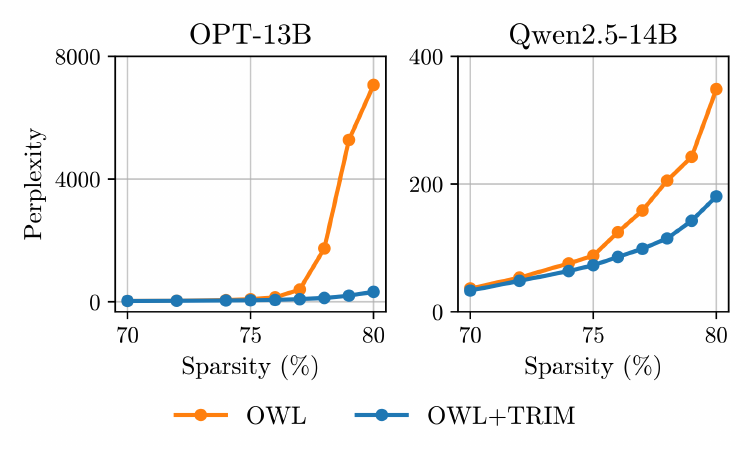}
    \caption{Perplexity progression from 70\% to 80\% sparsity. TRIM extends the usable sparsity range.}
    \label{fig:ppl_progression}
\end{figure}

\section{Experiments} \label{sec:experiments}
\paragraph{Models and Evaluation.} We evaluate TRIM on Qwen2.5-3B/-7B/-14B/-32B/-72B \citep{qwen2025qwen25technicalreport},
LLaMA-2-7B/-13B \citep{touvron2023llama2openfoundation}, and OPT-6.7B/-13B \citep{zhang2022optopenpretrainedtransformer}. 
Our evaluation adheres to established methodologies for pruning LLMs \citep{sun2024simpleeffectivepruningapproach, frantar2023sparsegptmassivelanguagemodels}. 
We calculate perplexity on the WikiText-2 test split \citep{merity2016pointersentinelmixturemodels} and assess zero-shot performance across a capability-oriented suite of 
tasks. We evaluate on BoolQ \citep{clark2019boolqexploringsurprisingdifficulty}, 
RTE \citep{wang2019gluemultitaskbenchmarkanalysis}, HellaSwag \citep{zellers2019hellaswagmachinereallyfinish}, WinoGrande \citep{sakaguchi2019winograndeadversarialwinogradschema}, 
ARC Easy and Challenge \citep{clark2018thinksolvedquestionanswering}, and OpenBookQA \citep{mihaylov2018suitarmorconductelectricity}, using the Language Model Evaluation Harness
\citep{eval-harness}. Additionally, we evaluate the generalizability of the WikiText-2 results on C4 and Pile in Appendix~\ref{sec:wikitext_generalization}.

\paragraph{Baselines.} We evaluate Wanda \citep{sun2024simpleeffectivepruningapproach}-based
OWL \citep{yin2024outlierweighedlayerwisesparsity} and AlphaPruning \citep{lu2024alphapruningusingheavytailedself} with and without TRIM.
In Appendix~\ref{sec:nakedWanda}, we evaluate Wanda with uniform layer-wise sparsities.
For all runs, we use the same number of calibration samples, each with a sequence length of 2048 tokens, 
randomly selected from the first shard of the C4 dataset \citep{raffel2023exploringlimitstransferlearning}. 
Hyperparameters can be found in Appendix~\ref{sec:Hyperparameters}.

\begin{table}[!b]
    \caption{Average zero-shot accuracies (\%) for models pruned at 70\% and 80\% sparsity. \(\Delta\) shows the average percentage increase achieved by adding TRIM. Higher is better.}
    \label{tab:mean_zero_shot_acc}
    \centering
    \begin{adjustbox}{width=0.98\textwidth, center}
    \begin{tabular}{clcccccccccc}
        \toprule
        \multirow{2.5}{*}{ } & \multirow{2.5}{*}{Method} & \multicolumn{2}{c}{OPT} & \multicolumn{2}{c}{LLaMA-2}  & \multicolumn{5}{c}{Qwen2.5} \\ 
        \cmidrule(lr){3-4}
        \cmidrule(lr){5-6}
        \cmidrule(lr){7-11}
         & & 6.7B & 13B & 7B & 13B & 3B & 7B & 14B & 32B & 72B & \(\Delta\) \\
        \midrule
        0\% & \multicolumn{1}{c}{-} & 51.52 & 52.60 & 59.73 & 63.02 & 61.13 & 65.82 & 68.12 & 68.09 & 70.12 & -\\
        \midrule
        \multirow{4}{*}{70\%}
        & Alpha      & 38.42 & 39.35 & \textbf{42.01} & 47.39 & 35.43 & 39.65 & 39.92 & 52.98 & 60.80 & 
        \multirow{2}{*}{\textit{+1.14}\%} \\
        & \cellcolor{gray!9}+TRIM & \cellcolor{gray!9}\textbf{42.27} & \cellcolor{gray!9}\textbf{42.27} & \cellcolor{gray!9}{41.60} & \cellcolor{gray!9}\textbf{47.92} & \cellcolor{gray!9}\textbf{36.75} & \cellcolor{gray!9}\textbf{39.75} & \cellcolor{gray!9}\textbf{40.73} & \cellcolor{gray!9}\textbf{53.15} & \cellcolor{gray!9}\textbf{61.76} \\
        \cmidrule(l){2-12}
        & OWL      & 40.67 & \textbf{43.60} & 43.30 & 49.01 & 36.53 & 39.22 & 40.92 & 54.55 & 61.30 &
        \multirow{2}{*}{\textit{+0.42}\%}\\ 
        & \cellcolor{gray!9}+TRIM  & \cellcolor{gray!9}\textbf{41.16} & \cellcolor{gray!9}43.12 & \cellcolor{gray!9}\textbf{43.36} & \cellcolor{gray!9}\textbf{49.54} & \cellcolor{gray!9}\textbf{36.65} & \cellcolor{gray!9}\textbf{39.82} & \cellcolor{gray!9}\textbf{41.68} & \cellcolor{gray!9}\textbf{55.04} & \cellcolor{gray!9}\textbf{62.47} \\
        \midrule
        \multirow{4}{*}{80\%}
        & Alpha     & 32.70 & 35.13 & 31.94 & 36.71 & 32.77 & 32.74 & \textbf{35.32} & 36.06 & 40.32 &
        \multirow{2}{*}{\textit{+0.65}\%}\\
        & \cellcolor{gray!9}+TRIM & \cellcolor{gray!9}\textbf{35.67} & \cellcolor{gray!9}\textbf{36.56} & \cellcolor{gray!9}\textbf{31.99} & \cellcolor{gray!9}\textbf{36.73} & \cellcolor{gray!9}\textbf{33.16} & \cellcolor{gray!9}\textbf{32.98} & \cellcolor{gray!9}34.40 & \cellcolor{gray!9}\textbf{37.44} & \cellcolor{gray!9}\textbf{40.59} \\
        \cmidrule(l){2-12}
        & OWL       & 32.61 & 35.22 & 32.60 & 36.23 & \textbf{35.42} & 32.31 & \textbf{33.70} & 37.35 & \textbf{41.12} &
        \multirow{2}{*}{\textit{+0.46}\%}\\
        &\cellcolor{gray!9}+TRIM   & \cellcolor{gray!9}\textbf{36.30} & \cellcolor{gray!9}\textbf{35.47} & \cellcolor{gray!9}\textbf{32.79} & \cellcolor{gray!9}\textbf{36.87} & \cellcolor{gray!9}34.82 & \cellcolor{gray!9}\textbf{32.64} & \cellcolor{gray!9}33.32 & \cellcolor{gray!9}\textbf{37.48} 
        & \cellcolor{gray!9}41.02 & \\ 
        \bottomrule
    \end{tabular}
    \end{adjustbox}
\end{table}

\subsection{Main results}
We evaluate OWL and AlphaPruning with and without TRIM at high sparsity ratios (70\% and 80\%), and additionally evaluate OWL with and without TRIM at 50\% and 60\%. 
For all methods, we only prune linear layers, excluding the initial embedding layer and the final classification head.

\paragraph{Language Modeling.} 

Table \ref{tab:perplexity_results} shows perplexity results for all models used in this study. Integrating TRIM with OWL and AlphaPruning consistently results in lower perplexity values than these baselines without TRIM. At 70\% and 80\% sparsity respectively, TRIM reduces perplexity for LLaMA-2-13B by 10\% and 31\%, and for Qwen2.5-14B by 9\% and 48\%. The effectiveness of TRIM is especially pronounced for OPT-13B at 80\% sparsity, where TRIM decreases OWL pruning perplexity from 6461.43 to 324.14.

TRIM improves the robustness and reliability of pruning results, correcting counterintuitive behaviors seen with baseline methods. At 80\% sparsity, Qwen2.5-14B shows higher perplexity than its smaller 7B counterpart, and OPT models suffer from unusually poor performance. TRIM resolves both issues, restoring expected scaling and stable performance.
These consistent and robust improvements across all model families, sizes, and sparsity levels underscore the critical role of effective dimension-wise sparsity allocation and highlight TRIM's efficacy in achieving it.

TRIM also mitigates two collapse patterns in the evaluated OWL configurations. At 80\% sparsity, baseline Qwen2.5-14B has higher perplexity than Qwen2.5-7B, and both OPT baselines exceed 6000 perplexity; TRIM restores the expected ordering for those two Qwen2.5 models and sharply reduces the OPT values, without restoring dense-model quality.

\paragraph{Zero-Shot Downstream Tasks.}
The benefits of TRIM extend beyond perplexity improvements to generalization on downstream tasks. 
As evidenced by the average zero-shot accuracies presented in Table~\ref{tab:mean_zero_shot_acc}, integrating TRIM with layer-wise pruning strategies yields superior performance compared to baseline methods without TRIM. This performance advantage holds across a wide range of models and sparsity levels.
The complete result tables can be found in Appendices \ref{tab:results_70} and \ref{tab:results_80}.

\subsection{Lower Sparsity Levels}\label{sec:lower_sparsity}
At moderate sparsity (50--60\%), TRIM matches or improves WikiText-2 validation perplexity relative to the OWL baseline across all tested models.
Table~\ref{tab:perplexity_results_low_sparsity} indicates that TRIM's benefits persist below the high-sparsity regime, enhancing or preserving quality at lower pruning ratios.

\begin{table}[H]
    \caption{Perplexity results on WikiText-2 for various models pruned at 50\% and 60\% sparsity. The \(\Delta\) column shows the average perplexity reduction across the tested models achieved by adding TRIM. Lower is better.}
    \label{tab:perplexity_results_low_sparsity}
    \centering
    \begin{adjustbox}{max width=0.85\textwidth, center}
    \begin{tabular}{clcccccccccc}
        \toprule
        \multirow{2.5}{*}{ } & \multirow{2.5}{*}{Method} & \multicolumn{2}{c}{OPT} & \multicolumn{2}{c}{LLaMA-2} & \multicolumn{5}{c}{Qwen2.5} \\
        \cmidrule(lr){3-4}
        \cmidrule(lr){5-6}
        \cmidrule(lr){7-11}
        & & 6.7B & 13B & 7B & 13B & 3B & 7B & 14B & 32B & 72B & \(\Delta\) \\
        \midrule
        \multirow{2}{*}{50\%}
        & OWL & 12.39 & 12.10 & 7.24 & 6.09 & 12.04 & 8.80 & 8.16 & \textbf{6.31} & 5.29 & \multirow{2}{*}{\textit{-0.5}\%}\\
        & \cellcolor{gray!9}+TRIM & \cellcolor{gray!9}12.39 & \cellcolor{gray!9}\textbf{12.08} & \cellcolor{gray!9}\textbf{7.15} & \cellcolor{gray!9}\textbf{5.99} & \cellcolor{gray!9}\textbf{11.94} & \cellcolor{gray!9}\textbf{8.79} & \cellcolor{gray!9}\textbf{8.12} & \cellcolor{gray!9}6.32 & \cellcolor{gray!9}\textbf{5.27} & \\
        \midrule
        \multirow{2}{*}{60\%}
        & OWL & 16.12 & 16.39 & 9.65 & 7.82 & 24.91 & 9.65 & 11.67 & 7.64 & 6.70 & \multirow{2}{*}{\textit{-2.3}\%} \\
        & \cellcolor{gray!9}+TRIM & \cellcolor{gray!9}\textbf{16.04} & \cellcolor{gray!9}\textbf{16.38} & \cellcolor{gray!9}\textbf{9.36} & \cellcolor{gray!9}\textbf{7.49} & \cellcolor{gray!9}\textbf{23.53} & \cellcolor{gray!9}\textbf{9.35} & \cellcolor{gray!9}\textbf{11.36} & \cellcolor{gray!9}\textbf{7.58} & \cellcolor{gray!9}\textbf{6.63} & \\
        \bottomrule
    \end{tabular}
    \end{adjustbox}
\end{table}

\subsection{Generalization to Other Pruning Metrics.}
Although we primarily evaluate TRIM with Wanda \citep{sun2024simpleeffectivepruningapproach}, its \emph{dimension-wise sparsity} and \emph{per-output} optimization apply to any scoring rule that
assigns a value to each weight. For clarity, the three metrics we test are:
(i) \textbf{Magnitude}, which uses absolute weights $|\mathbf{W}_{ij}|$ \cite{han2015learningweightsconnectionsefficient}; 
(ii) \textbf{SparseGPT}, which scores weights by the Hessian-aware ratio $[\lvert\mathbf{W}\rvert^2/\mathrm{diag}(\mathbf{H}^{-1})]_{ij}$ with $\mathbf{H}\!\approx\!\mathbf{X}^\top\mathbf{X}$ from layer inputs $\mathbf{X}$ \citep{frantar2023sparsegptmassivelanguagemodels}; 
(iii) \textbf{GBLM} which extends Wanda by multiplying weight magnitude with an activation- and gradient-based term, $|\mathbf{W}_{ij}|\!\cdot\!(\lVert\mathbf{X}_j\rVert+\alpha\,\mathbf{G}_{ij})$ (i.e., Wanda + gradient score) \citep{das2024sizegradientsshapepruning}.

SparseGPT compares candidates over blocks of 128 inputs (columns), Magnitude compares layer-wise, whereas Wanda/GBLM compare within a single output (row). 
For the SparseGPT row, we use its importance score but disable its post-pruning weight updates, isolating mask selection and allocation rather than evaluating full SparseGPT.
Table~\ref{tab:combined_sparsegpt_comparison} shows that TRIM's per-output configuration yields the best perplexity across all three metrics, indicating that TRIM gains generalize beyond the Wanda score.

\begin{table}[!b]
    \caption{WikiText-2 perplexity of LLaMA-2-13B at 70\% with OWL layer ratios. TRIM improves upon the per-output configuration and generalizes to other pruning metrics. The default comparison group is \underline{underlined}.}
    \label{tab:combined_sparsegpt_comparison}
    \centering
    \begin{adjustbox}{max width=\columnwidth, center}
    \begin{tabular}{lccc}
        \toprule
        Method & (layer) & (128, input) & (1, output) / +TRIM \\
        \midrule
         Magnitude & \underline{59.86} & 49.25 & 23.48 / \textbf{19.94} \\
         SparseGPT & 19.75 & \underline{19.23} & 15.19 / \textbf{14.11} \\
         GBLM & 18.53 & 22.80 &\underline{16.04} / \textbf{14.15} \\
        \bottomrule
    \end{tabular}
    \end{adjustbox}
\end{table}

\begin{figure}[!t]
  \centering

  \begin{subfigure}[b]{0.282\textwidth}
    \centering
    \raisebox{0mm}{\includegraphics[width=\linewidth]{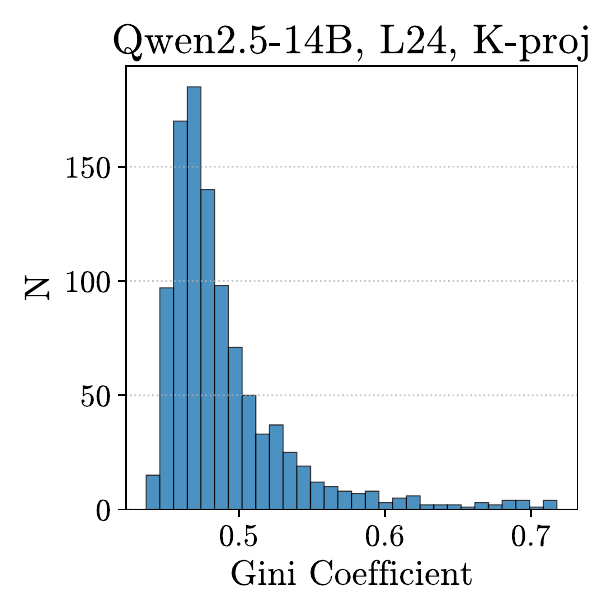}}
    \caption{Gini histogram.}
    \label{fig:gini_hist_middle}
  \end{subfigure}\hfill
  \begin{subfigure}[b]{0.28\textwidth}
    \centering
    \raisebox{1mm}{\includegraphics[width=\linewidth]{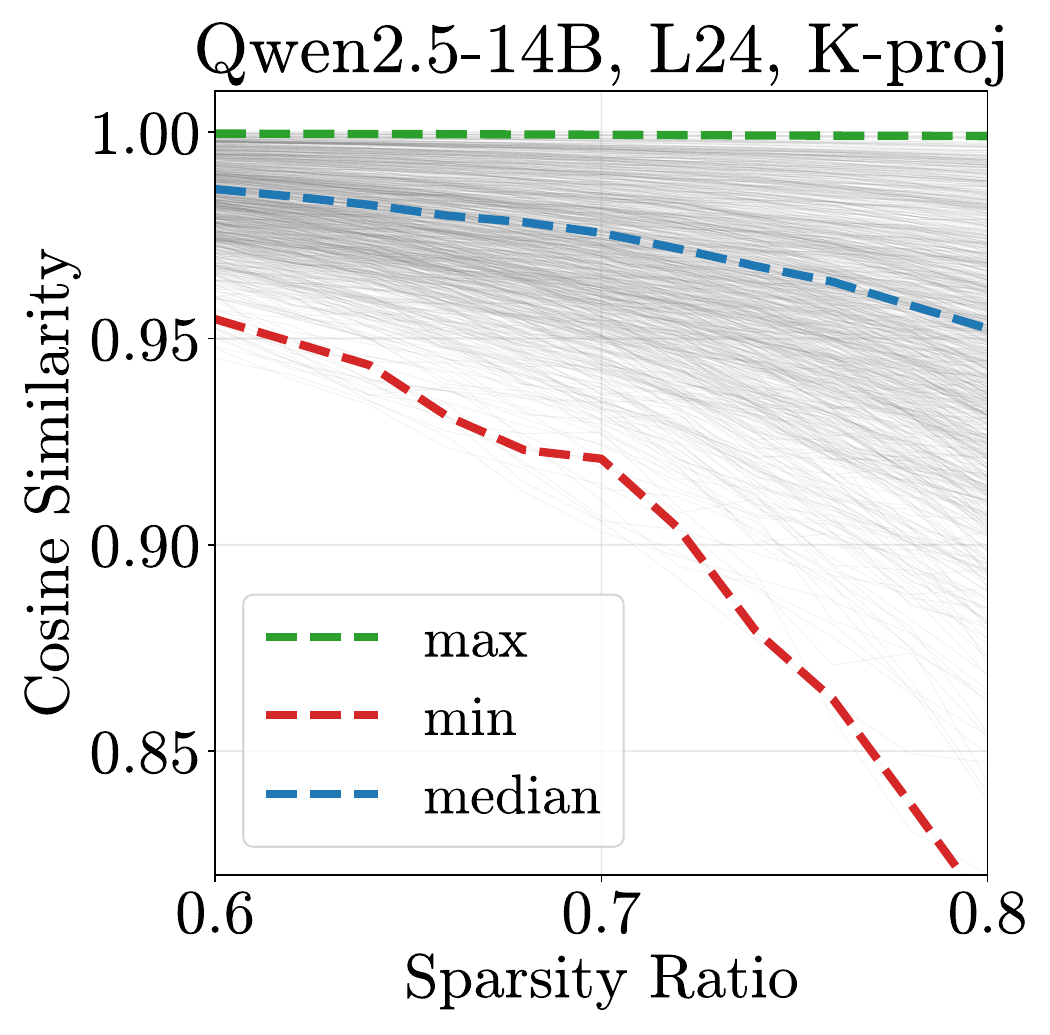}}
    \caption{Cosine similarity vs. sparsity.}
    \label{fig:cos_sparsity}
  \end{subfigure}\hfill
  \begin{subfigure}[b]{0.42\textwidth}
    \centering
    \raisebox{0.85mm}{\includegraphics[width=\linewidth]{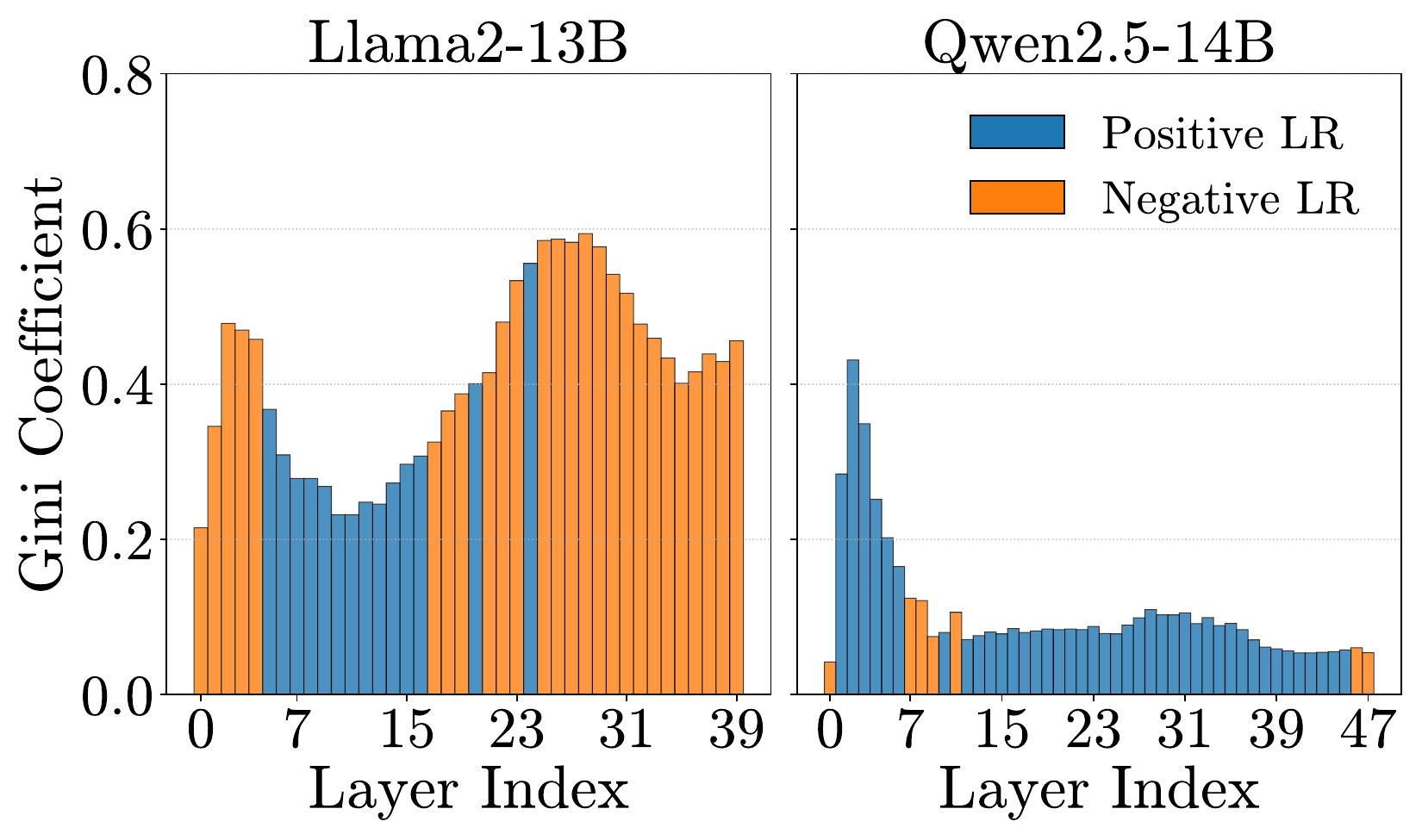}}
    \caption{LR sign vs Gini (outlier concentration).}
    \label{fig:gini_qwen_gate}
  \end{subfigure}

  \caption[Gallery of pruning diagnostics]{
    Gallery of pruning diagnostics.
    (a) \textbf{Gini histogram:} Gini coefficients of the Wanda pruning metric across output dimensions. Higher Gini $\Rightarrow$ signal concentrated in fewer weights. 
    (b) \textbf{Cosine similarity vs. sparsity:} Dimension-wise cosine similarities at increasing sparsity, showing heterogeneous pruning sensitivity across dimensions.
    (c) \textbf{LR sign vs outlier Gini:} LLaMA-2-13B (higher Gini; more concentrated outliers) uses more negative effective learning rates than Qwen2.5-14B (lower Gini; more uniform outliers). Both plots show the gate-projection.
  }
  \label{fig:gallery_all}
  \vspace{-1.15\baselineskip}
\end{figure}

\section{Analysis}\label{sec:Analysis}
In this section, we highlight three key empirical observations that explain TRIM's effectiveness: 1) output dimensions exhibit varying sensitivity to pruning; 2) post-pruning quality degrades as sparsity levels increase; and 3) some output dimensions are critical to model performance, while others contribute little to model performance.

\refstepcounter{observation}\paragraph{1. Output dimensions exhibit different sensitivity to pruning.}  
\label{obs:importance_concentration}
An output dimension's pruning sensitivity is closely tied to its importance score concentration. We examine the concentration of Wanda scores (\( |W_{i,:}| \cdot ||X||_2 \)) for each output dimension using Gini coefficients. Figure~\ref{fig:gini_hist_middle} displays histograms of these per-dimension Gini coefficients for the middle layer of Qwen2.5-14B. Figure~\ref{fig:gini_hist_middle} demonstrates that some dimensions rely on a few key weights while others distribute importance more broadly. These trends are consistent across models and model layers (see Appendix \ref{sec:MetricGiniTables}).

\refstepcounter{observation}\paragraph{2. Quality degradation accelerates with increasing sparsity.}\label{obs:accelerated_degradation}
Pruning can remove redundant weights with minimal impact at low sparsity levels. However, as sparsity rises, 
pruning impacts weights with essential activation signals, leading to a faster decay in performance. This non-linear decline is evident in the average trend of quality degradation on the dimension-level (Figure \ref{fig:cos_sparsity}) and 
in downstream metrics like the exponential increase in perplexity at high sparsity rates (Figure \ref{fig:ppl_progression}). 

The heterogeneity of importance concentration translates directly into differing pruning sensitivities across output dimensions. To demonstrate this, we increase the uniform sparsity ratio applied to a layer and measure the resulting quality degradation 
for \textit{each individual output dimension}. We quantify the quality using the cosine
similarity \( c_i = \text{CosineSimilarity}(Y_{i,:}, \hat{Y}_{i,:}) \) between the original output \( Y_{i,:} \) and the pruned output \( \hat{Y}_{i,:} \) for that dimension, given a set of sample inputs \(X\). Figure~\ref{fig:cos_sparsity} plots these cosine similarities \( c_i \) against increasing levels of sparsity \( T \). The rate at which quality 
degrades differs significantly across dimensions. Some dimensions maintain high similarity even at high sparsity levels (resilient to pruning), while others experience a rapid drop even at lower sparsity (sensitive to pruning). 

\refstepcounter{observation}\paragraph{3. Output dimensions differ in their impact on overall performance.}\label{obs:dimension_criticality}
Beyond differences in sensitivity, output dimensions can differ in their impact on model performance. As a diagnostic, we eliminate exactly one selected output dimension per layer in Qwen2.5-3B, resulting in 0.03\% total sparsity (Table~\ref{tab:norm_importance_ppl}). The selected row is the minimum- or maximum-L2-norm row, with a random-row control; this single setting has no uncertainty estimate and should not be generalized beyond the reported model.
Removing the dimension with the smallest L2-norm only increases perplexity by 0.16 from the unpruned baseline, similar to choosing a random dimension (+0.15). 
However, pruning the dimension with the highest norm increases perplexity to 273.10.

The importance of a dimension is also linked to the concentration of outlier positions present in it. Recall that $A_{ij}$ is an outlier position if $A_{ij} > M \cdot \bar{A}$. 
We take the top 10\% of dimensions with the largest number of outliers (i.e., "outlier-dense" dimensions) and prune them with 90\% sparsity, resulting in 9\% total model sparsity.
Note that pruning this way retains all outlier positions since they are always the last weights to get pruned.
Table~\ref{tab:outlier_rich_ppl} shows that pruning outlier-dense dimensions significantly impacts performance compared to randomly pruning an equivalent number of dimensions at the same sparsity.

\begin{table}[H]
    \centering
    \caption{Dimension-criticality diagnostics on WikiText-2 (perplexity; lower is better).}
    \label{tab:dimension_criticality_diagnostics}
    \begin{subtable}[t]{0.30\linewidth}
        \centering
        \caption{Removing one row per layer in Qwen2.5-3B.}
        \label{tab:norm_importance_ppl}
        \resizebox{\linewidth}{!}{%
        \begin{tabular}{lr}
          \toprule
          Method             & Perplexity \\
          \midrule
          Baseline           & 8.03 \\
          \(D\) w. min norm  & 8.19 \\
          \(D\) w. max norm  & \textbf{273.10} \\
          Random \(D\)       & 8.18 \\
          \bottomrule
        \end{tabular}}
    \end{subtable}\hfill
    \begin{subtable}[t]{0.66\linewidth}
        \centering
        \caption{Heavily pruning 10\% of rows (90\% row sparsity).}
        \label{tab:outlier_rich_ppl}
        \resizebox{\linewidth}{!}{%
        \begin{tabular}{lccc}
          \toprule
          Model        & Baseline & Random & Outlier-dense \(D\)\\
          \midrule
          Qwen2.5-7B   & 6.85     & 9.17   & \textbf{26.39} \\
          Qwen2.5-14B  & 5.29     & 7.70   & \textbf{26.16} \\
          LLaMA-2-7B   & 5.47     & 7.27   & \textbf{255.06} \\
          LLaMA-2-13B  & 4.88     & 6.23   & \textbf{74.27} \\
          \bottomrule
        \end{tabular}}
    \end{subtable}
\end{table}

First, outlier-dense dimensions are important for model performance.
Second, although outlier positions in dense dimensions appear to capture most of the signal, non-outlier weights retain substantial information, limiting aggressive pruning. This effect varies across model families. For example, LLaMA models exhibit a more pronounced degradation than Qwen, indicating that outlier-dense dimensions are more important for some models than for others.
Together, these diagnostics support the hypothesis that output dimensions are not uniformly important and that L2 norm or outlier statistics can identify unusually consequential rows in the tested settings.

\section{Discussion}\label{sec:discussion}
Targeted layer-wise allocation can improve pruning \citep{yin2024outlierweighedlayerwisesparsity,lu2024alphapruningusingheavytailedself,sun2025efficientshapleyvaluebasednonuniform,xu2025extremepruningllmsplugandplay}, and Lua-LLM independently learns allocations at both layer and intra-layer granularity \citep{lu2025luallm}. TRIM explores row-level allocation in a different regime: a forward-only wrapper around frozen weights and existing scorers. Our results show that this finer allocation can be especially useful at \textit{high sparsity levels}.
Observation \ref{obs:accelerated_degradation} demonstrates that post-pruning quality tends to degrade at an accelerated rate as sparsity increases. As a result, using coarse pruning strategies becomes increasingly detrimental. The core premise of TRIM is that significant heterogeneity exists at the dimension level (Observations \ref{obs:importance_concentration} and \ref{obs:dimension_criticality}). However, moving to such a fine-grained level of control introduces challenges: the search space for optimal 
sparsity configurations becomes substantially larger, and the feedback signals (e.g., quality metrics per dimension)
 might become less reliable or more noisy compared to layer-level aggregates.

TRIM addresses this complexity with two primary ideas. First, a positive learning rate pursues reduced variation in post-pruning quality degradation across dimensions, aiming to prevent any single dimension 
from experiencing significant quality loss.
Second, TRIM recognizes that simple quality equalization is not universally optimal. Some output dimensions possess both high initial post-pruning quality \textit{and} are critical to model function (e.g., outlier-dense dimensions; Table~\ref{tab:outlier_rich_ppl}). Forcing these dimensions to match the potentially lower quality of less important, more 
sensitive ones through aggressive variance reduction would be counterproductive.
TRIM can therefore use layer-quality feedback to select a negative learning rate. This reverses the update direction: rows with higher post-pruning quality receive lower sparsity targets, preserving more weights in them than positive quality equalization would. For outlier-dense dimensions, this lets TRIM favor rows whose concentrated signal yields high local quality but which remain disproportionately important to model performance (Table~\ref{tab:outlier_rich_ppl}). Consistent with this mechanism, negative rates are selected for approximately 40\% of LLaMA layers versus 10\% for the other tested models, coinciding with greater outlier concentration (Figure~\ref{fig:gini_qwen_gate}). The prevalence of this behavior may depend on the calibration data and rate search.

\section{Conclusion}
Our results show that uniform within-layer sparsity limits one-shot pruning quality. 
By redistributing each layer's sparsity budget across individual output rows, TRIM mitigates the collapse and 
instability that emerge at 70--80\% sparsity. For example, it reduces OPT-13B perplexity at 80\% sparsity 
from 6461.43 to 324.14. More broadly, dimension-wise sparsity demonstrates allocation as a valuable design axis, 
separate from the importance score used to rank weights.

This separation makes TRIM both practical and future-proof. It keeps model weights frozen, adds only seconds of allocation time, 
and can incorporate new importance scores without redesigning its 
allocation mechanism. As specialized kernels make high unstructured sparsity increasingly useful for real inference, lightweight 
allocation methods such as TRIM offer a practical way to reach that regime while preserving substantially more model quality.

\section*{Limitations}
Zero-shot changes are mixed, and the evaluated suite is not a reasoning-specific benchmark. 
Several 80\%-sparse models remain severely degraded despite large relative perplexity reductions, and would need
post pruning weight retraining to achieve dense-like performance.
The learning-rate search and negative-rate fallback may be sensitive to calibration data. 
TRIM redistributes a fixed per-matrix sparsity budget across rows without changing the matrix's overall sparsity or unstructured mask format. Systems such as Flash-LLM and SpInfer demonstrate that specialized kernels can exploit unstructured sparsity \citep{xia2023flashllm,fan2025spinfer}, but realized latency, throughput, energy, and memory gains depend on hardware and deployment configuration. We did not benchmark inference for TRIM masks and therefore do not claim such gains for TRIM.

\section*{LLM Usage Disclosure}
An LLM-based assistant was used to support manuscript organization, LaTeX editing, plotting code. 
All plotted values originate from the reported experiments; the authors manually reviewed and adjusted the resulting figures and take full responsibility for the manuscript and visualizations.

\bibliographystyle{colm2026_conference}
\bibliography{custom}

\appendix
\section{Quality Metric}\label{quality_metric}
TRIM utilizes two distinct quality evaluations to guide the pruning process. The first, layer-wise quality evaluation,
 assesses the pruning quality for an entire layer. 
We conduct an ablation study across different sparsity levels, model sizes, and model families, with results presented
 in Table \ref{tab:quality_metrics}. The metrics considered are Mean Squared Error (MSE), Cosine Flat 
 (which flattens the two-dimensional output vectors before calculating cosine similarity), and Cosine Sample 
 (which calculates cosine similarity along the sample axis and then averages the results). Our findings indicate 
 that both Cosine metrics offer a robust balance between accuracy and reliability across diverse pruning scenarios.

The second evaluation, dimension-wise quality assessment, measures the post-pruning quality for each individual output 
dimension. For this assessment, we investigate Cosine Similarity, Mean Squared Error (MSE), and Peak Signal-to-Noise 
Ratio (PSNR). As shown in Table \ref{tab:quality_metrics_dimwise}, while MSE can outperform Cosine Similarity in specific 
cases (e.g., for LLaMA models), Cosine Similarity generally demonstrates greater reliability.

These results demonstrate that: a) Cosine Similarity, as the default configuration for both layer-wise and dimension-wise 
quality assessments, proves to be a reliable choice, b) TRIM's performance can be further optimized in specific pruning 
scenarios by performing a hyperparameter sweep to identify the most suitable quality metric, and c) the choice of dimension-wise 
quality metric is typically more important than the layer-wise metric for overall pruning performance.

\begin{table}[H]
    \caption{Perplexity results for different layer quality metrics.}
    \label{tab:quality_metrics}
    \centering
    \begin{tabular}{lccccc}
        \toprule
        Model & Qwen2.5-3B & OPT-6.7B & OPT-6.7B & Qwen2.5-14B & LLaMA-2-13B \\
        Sparsity & 60\% & 70\% & 80\% & 80\% & 80\% \\
        \midrule
        Cosim Flat & \textbf{19.35} & 32.51 & 602.57 & 184.36 & 162.33 \\
        Cosim Sample & 19.38 & 32.75 & \textbf{590.12} & \textbf{170.23} & 173.09 \\
        MSE & 19.42 & \textbf{29.62} & 1970.02 & 189.66 & \textbf{142.81} \\
        \bottomrule
    \end{tabular}
\end{table}

\begin{table}[H]
    \caption{Perplexity results for different dimension-wise quality metrics.}
    \label{tab:quality_metrics_dimwise}
    \centering
    \begin{adjustbox}{max width=\linewidth}
    \begin{tabular}{lccccc}
        \toprule
        & Qwen2.5-3B & OPT-6.7B & OPT-6.7B& Qwen2.5-14B& LLaMA-2-13B\\
        Sparsity & 60\% & 70\% & 80\% & 80\% & 80\% \\
        \midrule
        Cosine Similarity & 19.35 & \textbf{32.51} & \textbf{602.57} & \textbf{184.36} & 162.34 \\
        PSNR              & 19.33 & 38.64         & 6834.34         & 216.87           & 226.88 \\
        MSE               & \textbf{19.26} & 44.72 & 4405.93 & 276.01 & \textbf{137.57} \\
        \bottomrule
    \end{tabular}
    \end{adjustbox}
\end{table}

\section{Runtime Overhead}\label{runtime_overhead}
One-shot pruning methods like Wanda \citep{sun2024simpleeffectivepruningapproach} are designed for efficient pruning of models 
with billions of parameters. Any extension to these methods should maintain this computational efficiency. 
TRIM, while iterative in nature, is designed around that and performs most computations efficiently vectorized on the GPU.

To empirically validate this claim, we prune the Qwen2.5-14B model using one NVIDIA A100-40GB and report the time in Table \ref{tab:runtime_overhead}.
TRIM adds little overhead compared to the duration of the entire Wanda based pruning function (8\% runtime increase for 32 samples, and 1.8\% increase for 512 samples).

\begin{table}[H]
    \caption{Runtime overhead of TRIM with different numbers of calibration samples.}
    \label{tab:runtime_overhead}
    \centering
    \begin{tabular}{lccccc}
        \toprule
        N samples & 32 & 64 & 128 & 256 & 512 \\
        \midrule
        WikiText-2 perplexity & 34.70 & 33.06 & 33.49 & 32.72 & 33.21 \\
        \midrule
        TRIM overhead (s) & 27.76 & 31.78 & 35.96 & 42.45 & 59.88 \\
        \midrule
        Total Time (s) & 375 & 562 & 888 & 1904 & 3325 \\
        \bottomrule
    \end{tabular}
\end{table}

\newpage
\section{TRIM without Layer-Wise Sparsity Allocation}\label{sec:nakedWanda}
In the main body of the paper, we evaluate TRIM in conjunction with layer-wise sparsity allocation schemes such as OWL \cite{yin2024outlierweighedlayerwisesparsity} and AlphaPruning \cite{lu2024alphapruningusingheavytailedself}.
To test if the improvements of TRIM hold up without these schemes, we evaluate several models on the Wanda metric with uniform sparsity ratios (see Table \ref{tab:perplexity_results_low_sparsity_nakedWanda}).
The improvements of TRIM \textit{without} layer-wise allocation schemes are comparable to the improvements \textit{with} layer-wise allocation schemes (compared to Table \ref{tab:perplexity_results_low_sparsity}). Because the baseline performance is typically worse without the layer-wise schemes, TRIM often shows larger relative gains.

\begin{table}[H]
    \caption{Perplexity results on WikiText-2 for various models pruned at 50\% and 60\% sparsity. The \(\Delta\) column shows the average perplexity reduction across the tested models achieved by adding TRIM. Lower is better.}
    \label{tab:perplexity_results_low_sparsity_nakedWanda}
    \centering
    \begin{adjustbox}{width=0.85\textwidth, center}
    \begin{tabular}{clcccccccc}
        \toprule
        \multirow{2.5}{*}{ } & \multirow{2.5}{*}{Method} & \multicolumn{1}{c}{OPT} & \multicolumn{2}{c}{LLaMA-2}  & \multicolumn{3}{c}{Qwen2.5} \\
        \cmidrule(lr){3-3}
        \cmidrule(lr){4-5}
        \cmidrule(lr){6-9}
        & & 6.7B & 7B & 13B & 3B & 7B & 14B & \(\Delta\) \\
        \midrule
        0\% & \multicolumn{1}{c}{-} & 10.86 & 5.47 & 4.88 & 8.03 & 6.85 & 5.29 & - \\
        \midrule
        \multirow{2}{*}{50\%}
        & Wanda & \textbf{11.96} & 6.92 & 5.98 & 11.37 & 8.57 & 7.30 & \multirow{2}{*}{\textit{-0.94}\%}\\
        & \cellcolor{gray!9}+TRIM & \cellcolor{gray!9}11.98 & \cellcolor{gray!9}\textbf{6.88} & \cellcolor{gray!9}\textbf{5.96} & \cellcolor{gray!9}\textbf{11.18} & \cellcolor{gray!9}\textbf{8.55} & \cellcolor{gray!9}\textbf{7.29} & \\
        \midrule
        \multirow{2}{*}{60\%}
        & Wanda & 15.14 & 10.77 & 8.41 & 21.51 & 13.75 & 11.31 & \multirow{2}{*}{\textit{-11.42}\%} \\
        & \cellcolor{gray!9}+TRIM & \cellcolor{gray!9}\textbf{14.68} & \cellcolor{gray!9}\textbf{10.64} & \cellcolor{gray!9}\textbf{8.26} & \cellcolor{gray!9}\textbf{19.06} & \cellcolor{gray!9}\textbf{13.05} & \cellcolor{gray!9}\textbf{9.98} & \\
        \bottomrule
    \end{tabular}
    \end{adjustbox}
\end{table}

\section{Robustness Analysis}\label{sec:robustness}
To evaluate the robustness, we run TRIM five times with different random seeds and report the results in Table \ref{tab:robustness}.
TRIM slightly increases the standard deviation compared to the OWL baseline method.

\begin{table}[H]
    \caption{Robustness analysis of TRIM with different random seeds on Qwen2.5-14B and LLaMA-2-7B at 70\% sparsity. WikiText-2 perplexity, lower is better.}
    \label{tab:robustness}
    \centering
    \begin{tabular}{lcccccccc}
        \toprule
        Model & Method & Seed 1 & Seed 2 & Seed 3 & Seed 4 & Seed 5 & Mean & Std \\
        \midrule
        Qwen2.5-14B & OWL & 37.63 & 37.43 & 36.33 & 37.71 & 36.87 & 37.20 & \textbf{0.52} \\
        Qwen2.5-14B & +TRIM & 33.64 & 33.50 & 33.61 & 33.16 & 34.95 & \textbf{33.78} & 0.61 \\
        \midrule
        LLaMA-2-7B & OWL & 21.53 & 21.39 & 21.32 & 21.58 & 21.60 & 21.48 & \textbf{0.11} \\
        LLaMA-2-7B & +TRIM & 20.44 & 20.37 & 19.95 & 20.40 & 20.73 & \textbf{20.38} & 0.25 \\
        \bottomrule
    \end{tabular}
\end{table}

\section{Implementation Details}
\paragraph{Sparsity Cutoff.} The limited amount of calibration data can push the sparsity for a dimension into a local minimum, sometimes
completely eliminating it. To prevent this overfitting from happening, we limit the sparsity that any dimension can experience to at most 95\%. 
This helps TRIM achieve more stable results.

\paragraph{Input Recalculation.} Additionally, we give the option for input vector recalculation.
One-shot pruning methods like Wanda prune block by block, and in each block layer by layer.
However, when the Q,K,V-projections are pruned the activation statistic for subsequent  layers (Out-projection,
or the MLP block) also changes. Therefore, we include an option to recalculate the input vectors used by TRIM.
Note that this \textit{does not} include the Wanda metric itself, which ensures fair comparisons. 
This recalculation makes pruning quality slightly better (see Table \ref{tab:recalc_effect}) and results in increased pruning time.
We recommend enabling this option for achieving the best results, and disabling it during hyperparameter tuning or developing.

\begin{table}[h]
    \caption{Effect of input recalculation on perplexity for Qwen2.5-14B at different sparsity levels.}
    \label{tab:recalc_effect}
    \centering
    \begin{adjustbox}{max width=\linewidth, center}
    \begin{tabular}{lccc}
        \toprule
        Sparsity & Baseline & Without Recalc & With Recalc \\
        \midrule
        70\% & 36.46 & 33.85 & 33.21 \\
        80\% & 348.48 & 186.32 & 180.67 \\
        \bottomrule
    \end{tabular}
  \end{adjustbox}
\end{table}
\newpage

\section{Generalizability of WikiText-2 Results}\label{sec:wikitext_generalization}
We expand our perplexity evaluations to showcase additional benchmark datasets beyond WikiText-2, specifically C4 and the Pile, consistent with prior work. The results are summarized in Tab~\ref{tab:wiki_generalization}.
\begin{table}[h]
\caption{Perplexity of dense, Wanda, and Wanda+TRIM models across datasets. Lower is better.}
\label{tab:wiki_generalization}
\centering
\small
\setlength{\tabcolsep}{5pt}
\begin{tabular}{llcc}
\toprule
Model & Dataset & Baseline & Wanda / +TRIM \\
\midrule
\multirow{3}{*}{\begin{tabular}[c]{@{}l@{}}Qwen2.5-7B\\(65\%)\end{tabular}} 
& WikiText-2 & 6.85 & 25.81 / \textbf{22.99} \\
& C4 & 11.88 & 34.64 / \textbf{31.15} \\
& Pile & 7.79 & 22.61 / \textbf{20.39} \\
\midrule
\multirow{3}{*}{\begin{tabular}[c]{@{}l@{}}LLaMA-2-7B\\(60\%)\end{tabular}} 
& WikiText-2 & 5.47 & 10.77 / \textbf{10.64} \\
& C4 & 7.26 & 13.94 / \textbf{13.80} \\
& Pile & 5.79 & 12.40 / \textbf{10.70} \\
\midrule
\multirow{3}{*}{\begin{tabular}[c]{@{}l@{}}LLaMA-2-13B\\(65\%)\end{tabular}} 
& WikiText-2 & 4.88 & 14.16 / \textbf{13.85} \\
& C4 & 6.73 & 19.32 / \textbf{18.98} \\
& Pile & 5.44 & 15.42 / \textbf{15.15} \\
\bottomrule
\end{tabular}
\end{table}\\
In these three settings, reductions in WikiText-2 perplexity are accompanied by reductions on C4 and the Pile, providing a limited cross-dataset check rather than establishing that WikiText-2 is universally predictive.

\section{Sensitivity Analysis of $K$}\label{sec:sensitivity_k}
The hyperparameter $K$ controls the number of iterations in the TRIM algorithm.
A larger $K$ allows for a more wide-ranging exploration of dimension-wise sparsity.
Our findings indicate that $K$ is a robust hyperparameter in these settings, with its impact depending
on the difficulty of the pruning task. In challenging scenarios (e.g., Qwen2.5-14B at 70\%), 
the iterative process is key, with performance improving significantly as k 
approaches our default of 10 before stabilizing. Conversely, in more straightforward scenarios 
(e.g., OPT-7B at 60\%), the model's performance is completely insensitive to k.
\\

\begin{table}[h]
    \caption{Effect of different values for hyperparameter $K$ in TRIM. Perplexity is reported on WikiText-2, averaged across two runs. Our default $K=10$ is \underline{underlined}.}
    \label{tab:k_sensitivity}
    \centering
    \begin{adjustbox}{max width=\linewidth, center}
    \begin{tabular}{lcc}
        \toprule
        $k$ & Qwen2.5-14B @ 70\% & OPT-7B @ 60\% \\
        \midrule
        3  & 99.46 & 14.73 \\
        6  & 66.70 & 14.73 \\
        \underline{10} & 56.67 & 14.73\\
        15 & 58.60 & 14.73 \\
        25 & 59.98 & 14.73 \\
        \bottomrule
    \end{tabular}
  \end{adjustbox}
\end{table}

\newpage
\section{Convergence Behavior}\label{sec:convergence_behavior}
To analyze convergence stability, we track the average sparsity shift per iteration. A value of 1\% means that the algorithm shifts all row sparsities by 1\% on average from iteration $N$ to iteration $N+1$.
Table~\ref{tab:convergence_behavior} reports the average shift for LLaMA-2-7B. The update magnitude peaks early ($N=3$ to $N=5$) and then steadily decreases, indicating that the largest allocation changes occur in the initial iterations before the distribution stabilizes.
\begin{table}[h]
    \caption{Average sparsity shift (\%) per iteration for LLaMA-2-7B at 50\% and 70\% sparsity.}
    \label{tab:convergence_behavior}
    \centering
    \begin{adjustbox}{max width=\linewidth, center}
    \begin{tabular}{lcc}
      \toprule
       & LLaMA-2-7B @ 50\% & LLaMA-2-7B @ 70\% \\
      \midrule
      $N=0$  & 0.26\% & 0.23\% \\
      $N=1$  & 0.27\% & 0.25\% \\
      $N=3$  & 0.27\% & 0.26\% \\
      $N=5$  & 0.25\% & 0.27\% \\
      $N=10$ & 0.21\% & 0.25\% \\
      $N=20$ & 0.17\% & 0.22\% \\
      $N=40$ & 0.10\% & 0.13\% \\
      \bottomrule
    \end{tabular}
    \end{adjustbox}
\end{table}

\newpage
\section{Hyperparameters}\label{sec:Hyperparameters}
\FloatBarrier
We publish the hyperparameters used for the layer-wise sparsity allocation methods.
For OWL \citep{yin2024outlierweighedlayerwisesparsity} we searched in the intervals $\lambda \in \{0.02, 0.05, 0.08, 0.12, 0.15, 0.2\}$ and $\mathbf{M} \in \{3, 5, 7, 10\}$.
For AlphaPruning \citep{lu2024alphapruningusingheavytailedself} we searched in the interval $\tau \in \{0.1, 0.2, 0.3, 0.4\}$.
All sweeps were done at 70\% sparsity with one seed and Wanda as the underlying \citep{sun2024simpleeffectivepruningapproach} pruning metric. 

\begin{table}[h]
    \caption{Hyperparameters for the layer-wise sparsity allocation methods that produce the results in this paper.}
    \label{tab:hyperparameters}
    \centering
    \begin{adjustbox}{max width=\linewidth, center}
    \begin{tabular}{lcc} 
      \toprule
      Model & OWL (M, $\lambda$) & AlphaPruning ($\tau$) \\ 
      \midrule
      OPT-6.7B & (10, 12\%) & 0.1 \\
      OPT-13B & (10, 5\%) & 0.1 \\
      LLaMA-2-7B & (5, 20\%) & 0.3 \\
      LLaMA-2-13B & (5, 15\%) & 0.3 \\
      Qwen2.5-3B & (5, 12\%) & 0.1 \\
      Qwen2.5-7B & (5, 12\%) & 0.1 \\
      Qwen2.5-14B & (3, 20\%) & 0.3 \\
      Qwen2.5-32B & (5, 15\%) & 0.3 \\
      Qwen2.5-72B &	(7, 8\%) & 0.1 \\
      \bottomrule
    \end{tabular}
    \end{adjustbox}
\end{table}

\onecolumn
\newpage
\section{Pruning Metric Distribution Plots}\label{sec:MetricGiniTables}
\begin{figure}[h]
    \centering
    \begin{adjustbox}{max width=0.88\linewidth}
        \includegraphics{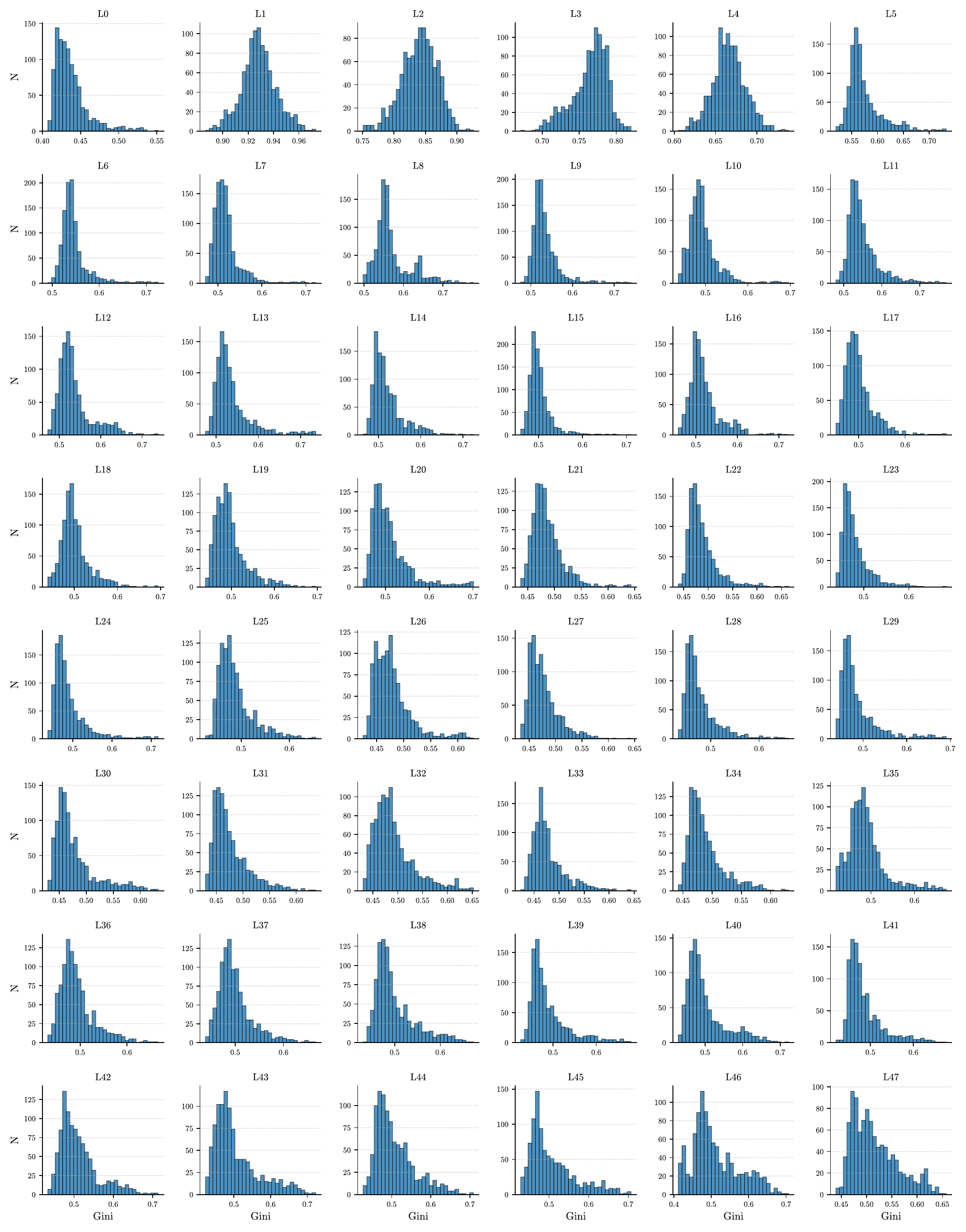}
    \end{adjustbox}
    \caption{Histogram showing the Gini coefficients for all layers of Qwen2.5-14B (K-proj). A high Gini means that the pruning metric (here: Wanda) is concentrated in fewer weights. }
    \label{fig:allGiniHists}
\end{figure}

We visualize the heterogeneity in the signal distribution at the dimension level in Fig \ref{fig:allGiniHists}.

\section{Licensing}
All used code, datasets and tasks are open-licensed under the MIT or CC by 4.0 license.

\section{Task Results}
Full 0-shot task result tables (\ref{tab:results_70}, \ref{tab:results_80}).
\begin{table*}[h]
    \caption{Accuracies (\%) for 7 zero-shot tasks for various models with unstructured 70\% sparsity and based on Wanda. Higher is better.
    We add TRIM to non-uniform layer-wise allocation methods OWL and AlphaPruning.}
    \label{tab:results_70}
    \centering
    \begin{adjustbox}{max width=\textwidth, max height=0.49\textheight, center}
    \begin{tabular}{cccccccccc}
        \toprule
        \multirow{2}{*}{Model} & \multirow{2}{*}{Method} & \multirow{2}{*}{BoolQ} & \multirow{2}{*}{RTE} & Hella- & Wino- & \multirow{2}{*}{ARC-e} & \multirow{2}{*}{ARC-c} & \multirow{2}{*}{OBQA} & \multirow{2}{*}{Mean} \\
        & & & & Swag & Grande & & & & \\
        \midrule
        \multirow{5}{*}{OPT-6.7B} & Dense & 66.09 & 55.23 & 50.48 & 65.19 & 65.61 & 30.46 & 27.60 & 51.52 \\
        \cmidrule(l){2-10}
        & Alpha & 62.14 & \textbf{55.96} & 28.99 & 52.49 & 40.07 & 17.06 & 12.20 & 38.42 \\
        & +TRIM & \textbf{62.17} & 53.43 & \textbf{35.72} & \textbf{54.54} & \textbf{48.86} & \textbf{22.35} & \textbf{18.80} & \textbf{42.27} \\ 
        \cmidrule(l){2-10}
        & OWL & 62.17 & 51.99 & 32.90 & \textbf{53.35} & 47.10 & 19.96 & 17.20 & 40.67 \\
        & +TRIM & \textbf{62.20} & \textbf{52.71} & \textbf{33.13} & 52.80 & \textbf{48.27} & \textbf{20.82} & \textbf{18.20} & \textbf{41.16} \\
        \midrule
        \multirow{5}{*}{OPT-13B} & Dense & 65.87 & 57.76 & 52.46 & 65.19 & 67.09 & 32.85 & 27.00 & 52.60 \\
        \cmidrule(l){2-10}
        & Alpha & 54.68 & \textbf{58.12} & 31.20 & 54.14 & 37.50 & 22.18 & \textbf{17.60} & 39.35 \\
        & +TRIM & \textbf{62.29} & 53.07 & \textbf{37.37} & \textbf{57.30} & \textbf{46.25} & \textbf{23.21} & 16.40 & \textbf{42.27} \\
        \cmidrule(l){2-10}
        & OWL & 62.23 & \textbf{57.40} & \textbf{37.65} & 58.01 & 47.35 & \textbf{25.34} & \textbf{17.20} & \textbf{43.60} \\
        & +TRIM & \textbf{62.26} & 54.51 & 37.33 & \textbf{58.56} & \textbf{48.23} & 23.72 & \textbf{17.20} & 43.12 \\
        \midrule
        \multirow{5}{*}{LLaMA-2-7B} & Dense & 77.71 & 62.82 & 57.19 & 69.22 & 76.35 & 43.43 & 31.40 & 59.73 \\
        \cmidrule(l){2-10}
        & Alpha & 62.26 & 52.71 & \textbf{34.74} & \textbf{61.64} & \textbf{43.22} & 22.10 & \textbf{17.40} & \textbf{42.01} \\
        & +TRIM & \textbf{62.45} & \textbf{53.07} & \textbf{34.74} & 58.56 & 42.97 & \textbf{22.18} & 17.20 & 41.60 \\
        \cmidrule(l){2-10}
        & OWL & 62.35 & \textbf{52.71} & 36.91 & \textbf{60.93} & \textbf{48.11} & 22.87 & 19.20 & 43.30 \\
        & +TRIM & \textbf{62.39} & \textbf{52.71} & \textbf{37.15} & 60.22 & 47.81 & \textbf{23.46} & \textbf{19.80} & \textbf{43.36} \\
        \midrule
        \multirow{5}{*}{LLaMA-2-13B} & Dense & 80.55 & 65.34 & 60.05 & 72.14 & 79.42 & 48.46 & 35.20 & 63.02 \\
        \cmidrule(l){2-10}
        & Alpha & 62.91 & 52.71 & 40.53 & \textbf{66.69} & 56.27 & \textbf{29.61} & \textbf{23.00} & 47.39 \\
        & +TRIM & \textbf{65.41} & \textbf{53.07} & \textbf{40.79} & 66.46 & \textbf{57.28} & 29.44 & \textbf{23.00} & \textbf{47.92} \\
        \cmidrule(l){2-10}
        & OWL & 70.43 & \textbf{52.71} & 41.27 & 65.90 & \textbf{58.96} & \textbf{29.61} & 24.20 & 49.01 \\
        & +TRIM & \textbf{71.62} & \textbf{52.71} & \textbf{41.82} & \textbf{67.48} & 58.75 & \textbf{29.61} & \textbf{24.80} & \textbf{49.54} \\
        \midrule
        \multirow{5}{*}{Qwen2.5-3B} & Dense & 77.34 & 75.45 & 54.95 & 68.35 & 77.36 & 44.88 & 29.60 & 61.13 \\
        \cmidrule(l){2-10}
        & Alpha & 53.91 & \textbf{53.43} & \textbf{27.72} & 50.04 & 33.33 & \textbf{18.17} & 11.40 & 35.43 \\
        & +TRIM & \textbf{61.71} & 52.71 & 27.68 & \textbf{50.20} & \textbf{34.47} & 17.66 & \textbf{12.80} & \textbf{36.75} \\
        \cmidrule(l){2-10}
        & OWL & \textbf{62.17} & \textbf{52.71} & 27.97 & \textbf{49.96} & 33.42 & \textbf{17.49} & \textbf{12.00} & 36.53 \\
        & +TRIM & \textbf{62.17} & \textbf{52.71} & \textbf{28.00} & 48.70 & \textbf{36.07} & 16.89 & \textbf{12.00} & \textbf{36.65} \\
        \midrule
        \multirow{5}{*}{Qwen2.5-7B} & Dense & 84.68 & 81.59 & 60.01 & 73.01 & 80.47 & 47.78 & 33.20 & 65.82 \\
        \cmidrule(l){2-10}
        & Alpha & \textbf{62.29} & \textbf{52.71} & 30.42 & \textbf{54.78} & 44.23 & 19.11 & 14.00 & 39.65 \\
        & +TRIM & 62.08 & \textbf{52.71} & \textbf{30.76} & 52.25 & \textbf{45.83} & \textbf{19.62} & \textbf{15.00 }& \textbf{39.75} \\
        \cmidrule(l){2-10}
        & OWL & 62.14 & \textbf{52.71} & 30.35 & 52.33 & 45.08 & 17.92 & \textbf{14.00} & 39.22 \\
        & +TRIM & \textbf{62.20} & \textbf{52.71} & \textbf{30.63} & \textbf{53.75} & \textbf{45.92} & \textbf{19.54} & \textbf{14.00} & \textbf{39.82} \\
        \midrule
        \multirow{5}{*}{Qwen2.5-14B} & Dense & 85.23 & 80.14 & 63.36 & 75.30 & 82.45 & 55.97 & 34.40 & 68.12 \\
        \cmidrule(l){2-10}
        & Alpha & 62.17 & \textbf{52.71} & 29.86 & 54.46 & 47.09 & 18.94 & 14.20 & 39.92 \\
        & +TRIM & \textbf{62.20} & \textbf{52.71} & \textbf{30.66} & \textbf{55.56} & \textbf{48.57} & \textbf{20.22} & \textbf{15.20} & \textbf{40.73} \\
        \cmidrule(l){2-10}
        & OWL & \textbf{62.17} & 52.07 & 31.82 & 57.14 & 48.36 & 19.28 & \textbf{15.60} & 40.92 \\
        & +TRIM & \textbf{62.17} & \textbf{52.71} & \textbf{32.54} & \textbf{58.88} & \textbf{50.42} & \textbf{19.45} & \textbf{15.60} & \textbf{41.68} \\
        \midrule
        \multirow{5}{*}{Qwen2.5-32B} & Dense & 87.12 & 81.59 & 64.98 & 75.22 & 80.85 & 53.07 & 33.80 & 68.09 \\
        \cmidrule(l){2-10}
        & Alpha & 68.93 & \textbf{64.98} & 43.43 & \textbf{69.69} & 66.58 & 33.02 & 24.20 & 52.98 \\
        & +TRIM & \textbf{70.24} & 62.09 & \textbf{44.19} & 69.38 & \textbf{67.55} & \textbf{33.79} & \textbf{24.80} & \textbf{53.15} \\
        \cmidrule(l){2-10}
        & OWL & 71.38 & \textbf{62.81} & 44.16 & 70.09 & 71.21 & 36.43 & \textbf{25.80} & 54.55 \\
        & +TRIM & \textbf{72.57} & 62.45 & \textbf{45.15} & \textbf{70.64} & \textbf{72.35} & \textbf{36.95} & 25.20 & \textbf{55.04} \\
        \midrule
        \multirow{5}{*}{Qwen2.5-72B} & Dense & 89.24 & 77.26 & 67.58 & 77.58 & 85.06 & 58.36 & 35.80 & 70.12 \\
        \cmidrule(l){2-10}
        & Alpha & 77.43 & 79.06 & 49.83 & 74.35 & 74.12 & 43.00 & 27.80 & 60.80 \\
        & +TRIM & \textbf{80.31} & \textbf{80.87} & \textbf{50.09} & \textbf{74.59} & \textbf{74.66} & \textbf{43.17} & \textbf{28.60} & \textbf{61.76} \\
        \cmidrule(l){2-10}
        & OWL & 76.61 & 78.34 & 50.77 & 74.43 & \textbf{76.26} & \textbf{44.88} & 27.80 & 61.30 \\
        & +TRIM & \textbf{81.80} & \textbf{80.50} & \textbf{50.92} & \textbf{75.06} & \textbf{76.26} & 44.11 & \textbf{28.60} & \textbf{62.47} \\
        \bottomrule
    \end{tabular}
\end{adjustbox}
\end{table*}

\begin{table*}[t]
    \caption{Accuracies (\%) for 7 zero-shot tasks for various models with unstructured 80\% sparsity and based on Wanda. Higher is better.
    We add TRIM to non-uniform layer-wise allocation methods OWL and AlphaPruning.}
    \label{tab:results_80}
    \centering
    \begin{adjustbox}{max width=\textwidth, max height=0.49\textheight, center}
        \begin{tabular}{cccccccccc}
            \toprule
            \multirow{2}{*}{Model} & \multirow{2}{*}{Method} & \multirow{2}{*}{BoolQ} & \multirow{2}{*}{RTE} & Hella- & Wino- & \multirow{2}{*}{ARC-e} & \multirow{2}{*}{ARC-c} & \multirow{2}{*}{OBQA} & \multirow{2}{*}{Mean} \\
            & & & & Swag & Grande & & & & \\
            \midrule
            \multirow{5}{*}{OPT-6.7B} & Dense & 66.09 & 55.23 & 50.48 & 65.19 & 65.61 & 30.46 & 27.60 & 51.52 \\
            \cmidrule(l){2-10}
            & Alpha & 41.04 & \textbf{52.71} & 25.93 & 50.20 & 27.44 & \textbf{19.97} & \textbf{11.60} & 32.70 \\
            & +TRIM & \textbf{60.70} & 50.18 & \textbf{26.57} & \textbf{51.38} & \textbf{31.19} & 18.69 & 11.00 & \textbf{35.67} \\
            \cmidrule(l){2-10}
            & OWL & 38.62 & 51.99 & 25.90 & 50.75 & 25.97 & \textbf{20.65} & \textbf{14.40} & 32.61 \\
            & +TRIM & \textbf{61.93} & \textbf{52.71} & \textbf{26.86} & \textbf{51.38} & \textbf{33.00} & 17.41 & 10.80 & \textbf{36.30} \\
            \midrule
            \multirow{5}{*}{OPT-13B} & Dense & 65.87 & 57.76 & 52.46 & 65.19 & 67.09 & 32.85 & 27.00 & 52.60 \\
            \cmidrule(l){2-10}
            & Alpha & 60.83 & 51.26 & 25.83 & 49.49 & 26.94 & \textbf{19.96} & 11.60 & 35.13 \\
            & +TRIM & \textbf{61.87} & \textbf{52.71} & \textbf{26.75} & \textbf{52.25} & \textbf{29.80} & 19.54 & \textbf{13.00} & \textbf{36.56} \\
            \cmidrule(l){2-10}
            & OWL & 56.39 & \textbf{52.71} & 26.10 & \textbf{51.14} & 26.89 & \textbf{20.90} & \textbf{12.40} & 35.22 \\
            & +TRIM & \textbf{58.01} & \textbf{52.71} & \textbf{26.99} & 50.67 & \textbf{30.35} & 17.75 & 11.80 & \textbf{35.47} \\
            \midrule
            \multirow{5}{*}{LLaMA-2-7B} & Dense & 77.71 & 62.82 & 57.19 & 69.22 & 76.35 & 43.43 & 31.40 & 59.73 \\
            \cmidrule(l){2-10}
            & Alpha & \textbf{37.89} & \textbf{52.71} & \textbf{26.20} & \textbf{47.28} & \textbf{26.73} & 20.14 & 12.60 & 31.94 \\
            & +TRIM & \textbf{37.89} & \textbf{52.71} & 26.18 & 47.04 & 26.56 & \textbf{20.56} & \textbf{13.00} & \textbf{31.99} \\
            \cmidrule(l){2-10}
            & OWL & 38.19 & \textbf{52.71} & \textbf{26.64} & 50.36 & 27.82 & 18.69 & \textbf{13.80} & 32.60 \\
            & +TRIM & \textbf{38.26} & \textbf{52.71} & 26.50 & \textbf{51.38} & \textbf{27.86} & \textbf{19.45} & 13.40 & \textbf{32.79} \\
            \midrule
            \multirow{5}{*}{LLaMA-2-13B} & Dense & 80.55 & 65.34 & 60.05 & 72.14 & 79.42 & 48.46 & 35.20 & 63.02 \\
            \cmidrule(l){2-10}
            & Alpha & 62.05 & \textbf{52.71} & 27.74 & \textbf{52.72} & 30.68 & 19.28 & 11.80 & 36.71 \\
            & +TRIM & \textbf{62.17} & \textbf{52.71} & \textbf{27.90} & 51.62 & \textbf{30.85} & \textbf{19.88} & \textbf{12.00} & \textbf{36.73} \\
            \cmidrule(l){2-10}
            & OWL & 61.41 & \textbf{52.71} & 27.49 & 50.75 & 30.35 & 19.11 & 11.80 & 36.23 \\
            & +TRIM & \textbf{62.08} & \textbf{52.71} & \textbf{28.06} & \textbf{53.04} & \textbf{30.64} & \textbf{19.37} & \textbf{12.20} & \textbf{36.87} \\
            \midrule
            \multirow{5}{*}{Qwen2.5-3B} & Dense & 77.34 & 75.45 & 54.95 & 68.35 & 77.36 & 44.88 & 29.60 & 61.13 \\
            \cmidrule(l){2-10}
            & Alpha & 37.77 & \textbf{55.23} & 26.12 & \textbf{50.04} & 27.06 & 19.80 & 13.40 & 32.77 \\
            & +TRIM & \textbf{44.28} & 49.46 & \textbf{26.47} & 48.54 & \textbf{28.58} & \textbf{20.56} & \textbf{14.20} & \textbf{33.16} \\
            \cmidrule(l){2-10}
            & OWL & \textbf{56.73} & \textbf{52.71} & 26.78 & \textbf{51.22} & \textbf{30.30} & \textbf{19.03} & 11.20 & \textbf{35.42} \\
            & +TRIM & 54.10 & \textbf{52.71} & \textbf{26.81} & 49.25 & 29.71 & 18.17 & \textbf{13.00} & 34.82 \\
            \midrule
            \multirow{5}{*}{Qwen2.5-7B} & Dense & 84.68 & 81.59 & 60.01 & 73.01 & 80.47 & 47.78 & 33.20 & 65.82 \\
            \cmidrule(l){2-10}
            & Alpha & \textbf{37.92} & 52.71 & 26.91 & 49.88 & 28.37 & \textbf{19.03} & \textbf{14.40} & 32.74 \\
            & +TRIM & 37.77 & \textbf{53.06} & \textbf{27.21} & \textbf{49.96} & \textbf{30.01} & 18.86 & 14.00 & \textbf{32.98} \\
            \cmidrule(l){2-10}
            & OWL & 37.89 & \textbf{52.71} & 27.14 & 47.28 & \textbf{30.51} & 17.83 & 12.80 & 32.31 \\
            & +TRIM & \textbf{38.29} & \textbf{52.71} & \textbf{27.30} & \textbf{47.83} & 30.09 & \textbf{18.09} & \textbf{14.20} & \textbf{32.64} \\
            \midrule
            \multirow{5}{*}{Qwen2.5-14B} & Dense & 85.23 & 80.14 & 63.36 & 75.30 & 82.45 & 55.97 & 34.40 & 68.12 \\
            \cmidrule(l){2-10}
            & Alpha & \textbf{59.17} & \textbf{52.71} & \textbf{26.94} & 48.93 & 30.43 & \textbf{16.64} & 12.40 & \textbf{35.32} \\
            & +TRIM & 50.67 & \textbf{52.71} & 26.86 & \textbf{50.12} & \textbf{30.89} & 16.55 & \textbf{13.00} & 34.40 \\
            \cmidrule(l){2-10}
            & OWL & \textbf{44.59} & \textbf{52.71} & \textbf{27.27} & 49.17 & 31.94 & 16.64 & \textbf{13.60} & \textbf{33.70} \\
            & +TRIM & 42.08 & \textbf{52.71} & 27.08 & \textbf{49.25} & \textbf{32.15} & \textbf{16.98} & 13.00 & 33.32 \\
            \midrule
            \multirow{5}{*}{Qwen2.5-32B} & Dense & 87.12 & 81.59 & 64.98 & 75.22 & 80.85 & 53.07 & 33.80 & 68.09 \\
            \cmidrule(l){2-10}
            & Alpha & 59.14 & 52.71 & 27.61 & 48.22 & 35.19 & 17.15 & 12.40 & 36.06 \\
            & +TRIM & \textbf{61.83} & \textbf{53.43} & \textbf{28.79} & \textbf{49.49} & \textbf{37.46} & \textbf{17.49} & \textbf{13.60} & \textbf{37.44} \\
            \cmidrule(l){2-10}
            & OWL & 62.08 & \textbf{52.71} & \textbf{28.49} & 49.57 & 38.38 & \textbf{17.24} & 13.00 & 37.35 \\
            & +TRIM & \textbf{62.20} & \textbf{52.71} & 28.35 & \textbf{50.20} & \textbf{39.18} & 16.30 & \textbf{13.40} & \textbf{37.48} \\
            \midrule
            \multirow{5}{*}{Qwen2.5-72B} & Dense & 89.24 & 77.26 & 67.58 & 77.58 & 85.06 & 58.36 & 35.80 & 70.12 \\
            \cmidrule(l){2-10}
            & Alpha & \textbf{62.17} & \textbf{52.71} & 32.14 & 58.41 & 41.12 & \textbf{21.50} & 14.20 & 40.32 \\
            & +TRIM & \textbf{62.17} & \textbf{52.71} & \textbf{32.50} & \textbf{58.72} & \textbf{41.24} & 21.16 & \textbf{15.60} & \textbf{40.59} \\
            \cmidrule(l){2-10}
            & OWL & \textbf{62.17} & \textbf{52.71} & 32.63 & \textbf{58.88} & 43.56 & \textbf{21.67} & \textbf{16.20} & \textbf{41.12} \\
            & +TRIM & \textbf{62.17} & \textbf{52.71} & \textbf{32.69} & 57.85 & \textbf{44.28} & 21.42 & 16.00 & 41.02 \\
            \bottomrule
        \end{tabular}
\end{adjustbox}
\end{table*}

\FloatBarrier

\end{document}